\documentclass[manuscript,screen]{acmart}

\AtBeginDocument{%
  \providecommand\BibTeX{{%
    \normalfont B\kern-0.5em{\scshape i\kern-0.25em b}\kern-0.8em\TeX}}}

\setcopyright{acmcopyright}
\acmJournal{TOMM}
\acmYear{2021} \acmVolume{1} \acmNumber{1} \acmArticle{1} \acmMonth{1} \acmPrice{15.00}\acmDOI{10.1145/3472623}

\usepackage{amsmath}
\usepackage{amsfonts}
\usepackage{array, multirow, boldline}
\usepackage{subfigure}
\usepackage{algorithm}
\usepackage{algpseudocode}
\usepackage{xcolor}
\usepackage{hyperref}
\usepackage{url}
\usepackage{comment}
\newcommand\tab[1][0.2cm]{\hspace*{#1}}

\begin{document}
\title{Mask or Non-Mask? Robust Face Mask Detector via Triplet-Consistency Representation Learning}


\author{Chun-Wei Yang}
\affiliation{
  \institution{National Yang Ming Chiao Tung University}
  \streetaddress{1001 University Road}
  \city{Hsinchu}
  \country{Taiwan}}
\email{wei1997.ee08g@nctu.edu.tw}

\author{Thanh Hai Phung}
\affiliation{
  \institution{National Yang Ming Chiao Tung University}
  \streetaddress{1001 University Road}
  \city{Hsinchu}
  \country{Taiwan}}
\email{haipt.eed08g@nctu.edu.tw}

\author{Hong-Han Shuai}
\affiliation{
  \institution{National Yang Ming Chiao Tung University}
  \streetaddress{1001 University Road}
  \city{Hsinchu}
  \country{Taiwan}}
\email{hhshuai@nctu.edu.tw}

\author{Wen-Huang Cheng}
\affiliation{
  \institution{National Yang Ming Chiao Tung University and National Chung Hsing University}
  \streetaddress{1001 University Road}
  \city{Hsinchu}
  \country{Taiwan}}
\email{whcheng@nctu.edu.tw}

\renewcommand{\shortauthors}{Yang et al.}

\begin{abstract}
In the absence of vaccines or medicines to stop COVID-19, one of the effective methods to slow the spread of the coronavirus and reduce the overloading of healthcare is to wear a face mask. Nevertheless, to mandate the use of face masks or coverings in public areas, additional human resources are required, which is tedious and attention-intensive. To automate the monitoring process, one of the promising solutions is to leverage existing object detection models to detect the faces with or without masks. As such, security officers do not have to stare at the monitoring devices or crowds, and only have to deal with the alerts triggered by the detection of faces without masks. Existing object detection models usually focus on designing the CNN-based network architectures for extracting discriminative features. However, the size of training datasets of face mask detection is small, while the difference between faces with and without masks is subtle. Therefore, in this paper, we propose a face mask detection framework that uses the context attention module to enable the effective attention of the feed-forward convolution neural network by adapting their attention maps feature refinement. Moreover, we further propose an anchor-free detector with Triplet-Consistency Representation Learning by integrating the consistency loss and the triplet loss to deal with the small-scale training data and the similarity between masks and occlusions. Extensive experimental results show that our method outperforms the other state-of-the-art methods. The source code is released as a public download to improve public health at \url{https://github.com/wei-1006/MaskFaceDetection}.
\end{abstract}



\maketitle
\section{Introduction}
As the saying goes, ``An ounce of prevention is worth a pound of cure.'' To alleviate the burden of healthcare caused by COVID-19, World Health Organization (WHO) suggests that masks should be used as part of a comprehensive strategy of measures to suppress transmission and provide protection against COVID-19. Leung et al.~\cite{leung2020respiratory} also show that surgical masks can protect people from the coronavirus by reducing the probability of airborne transmission. As such, many governments mandate the use of face masks or coverings in public areas, such as retail establishments and public transportation~\cite{borkar2017real,jian2010face}. Nevertheless, to implement the executive order, it requires staff to monitor whether pedestrians wear masks or not at the entrance of restricted areas, which may be a heavy burden for large-scale and long-term surveillance.

\begin{figure}[t]
\centering
		\subfigure[Faces of different orientations]{\label{fig:fig1_orientation}
			\includegraphics[width = 0.3\linewidth, height = 3cm]{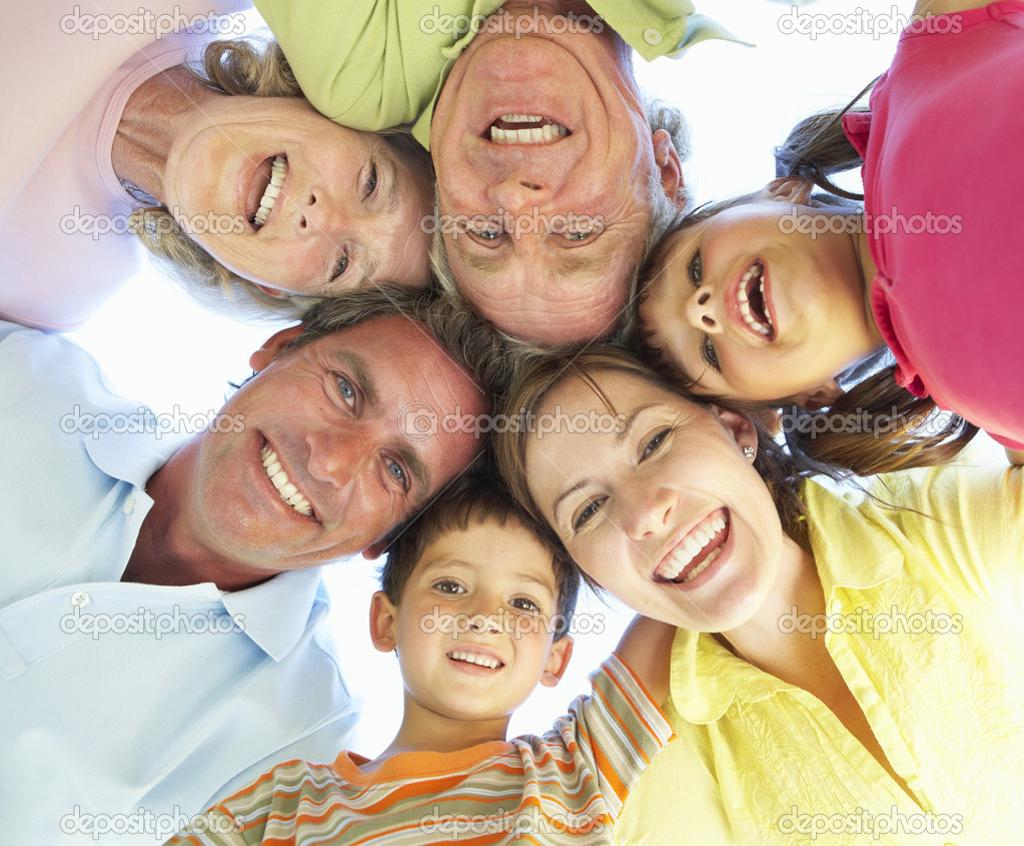}
		}
		~
		\subfigure[Reflected faces without masks]{\label{fig:fig1_Reflected}
			\includegraphics[width = 0.3\linewidth, height = 3cm]{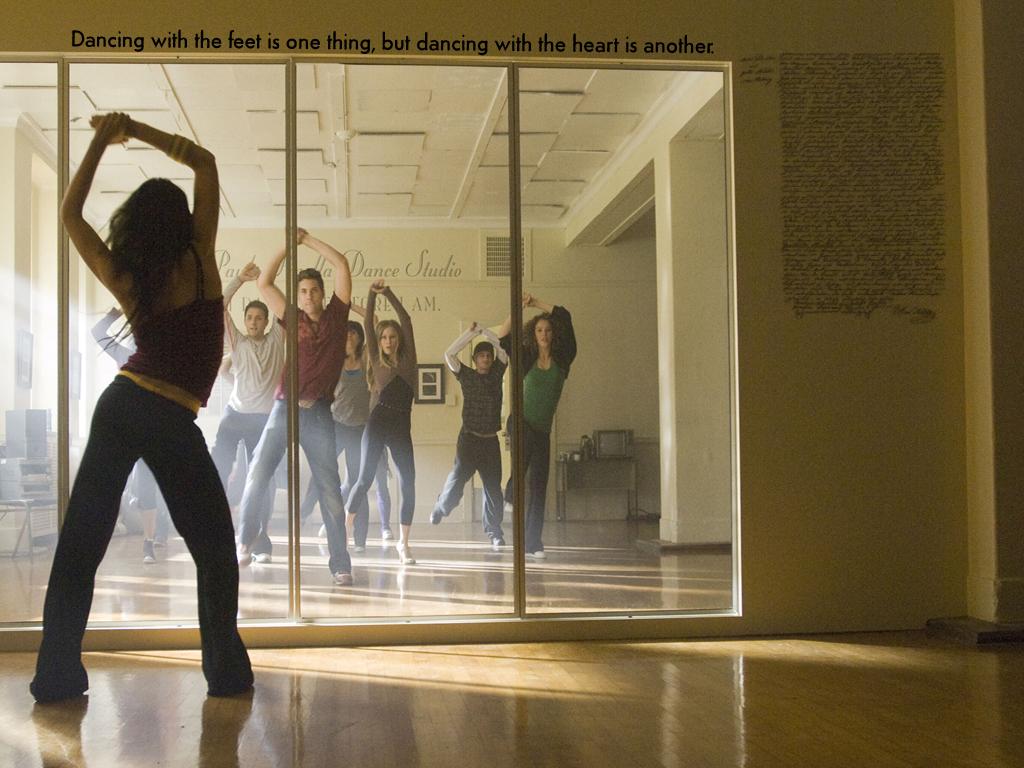}
		}
		~
		\subfigure[Crowd scene with different scales]{\label{fig:fig1_Crowded}
			\includegraphics[width = 0.3\linewidth, height = 3cm]{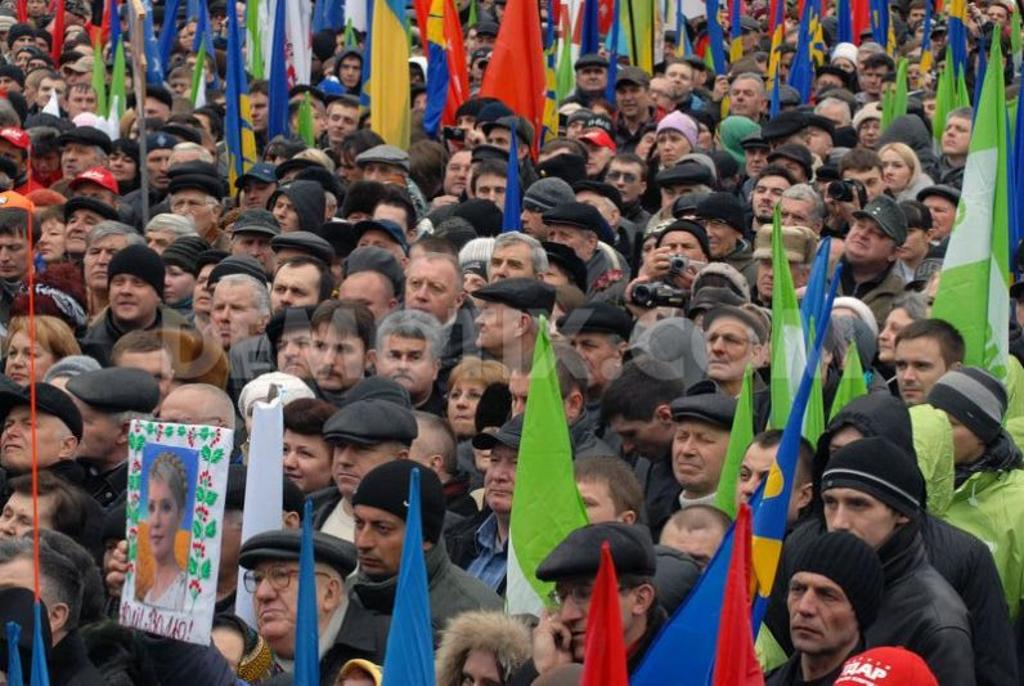}
		}\\
		~
		\subfigure[Occlusion and masks in different colors]{\label{fig:fig1_differentcolors}
			\includegraphics[width = 0.3\linewidth, height = 3cm]{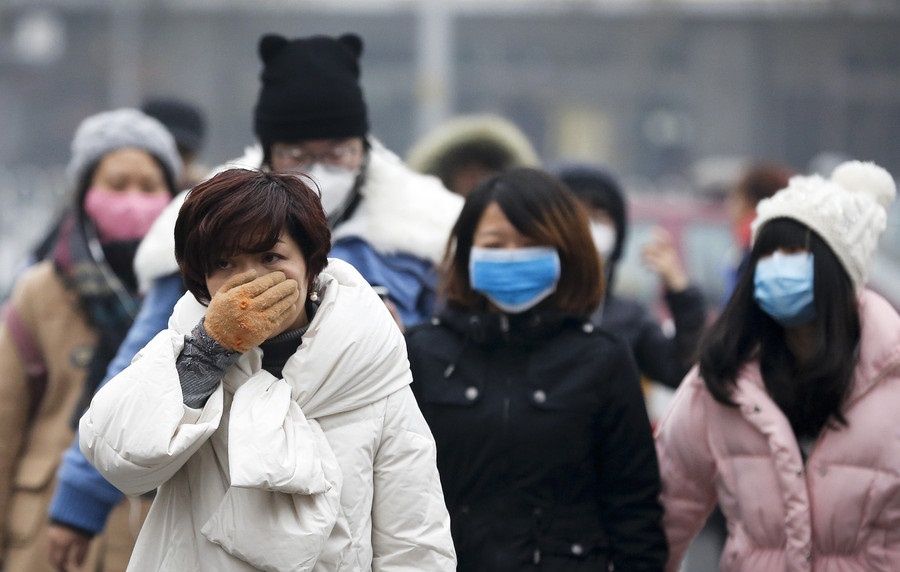}
		}
		~
		\subfigure[Faces with different poses]{\label{fig:fig1_poses}
			\includegraphics[width = 0.3\linewidth, height = 3cm]{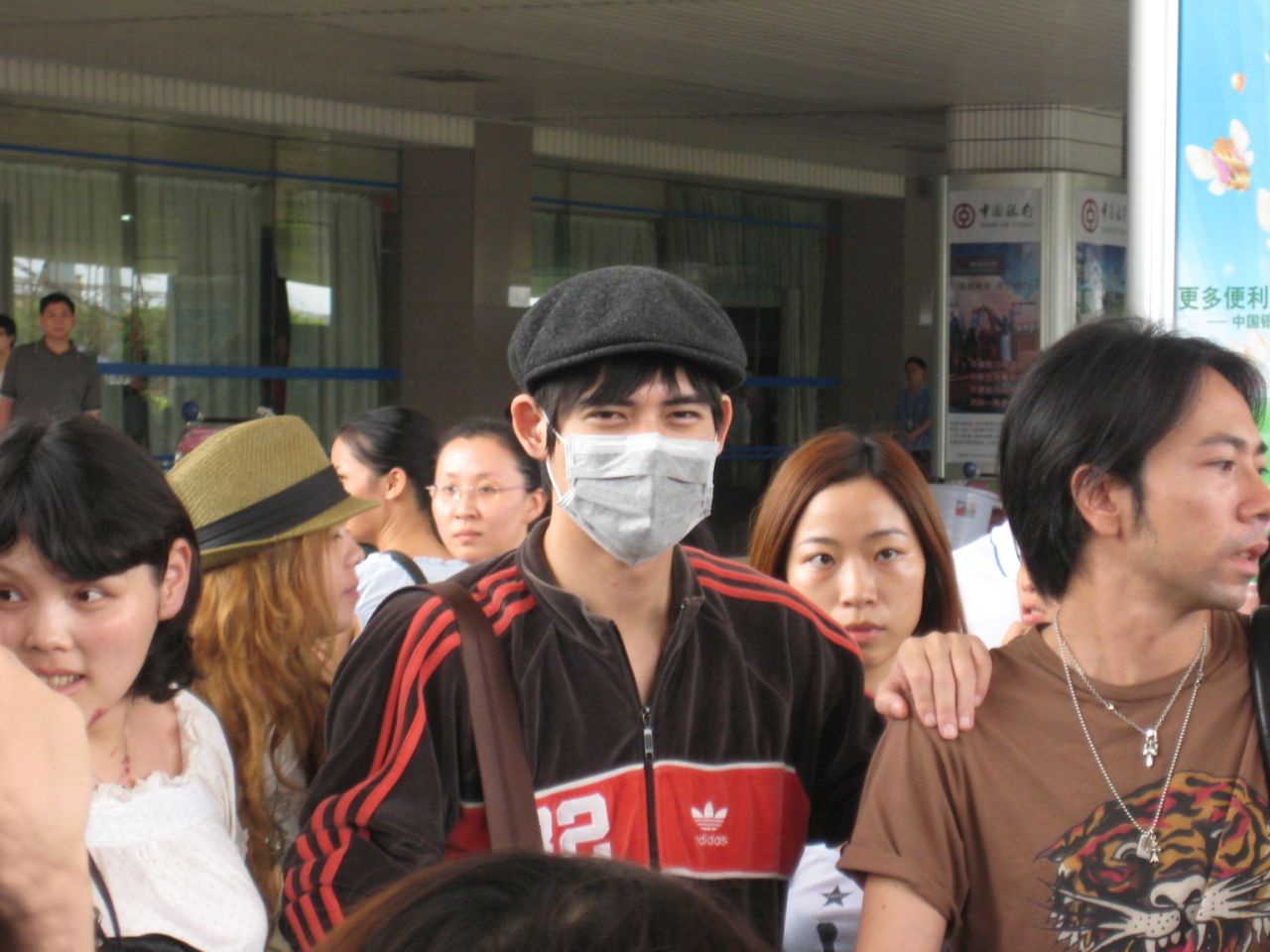}
		}
		~
		\subfigure[Masks on non-human]{\label{fig:fig1_nonhuman}
			\includegraphics[width = 0.3\linewidth, height = 3cm]{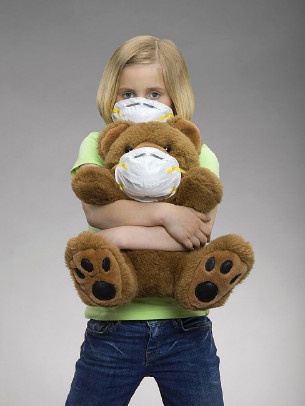}
		}\\
		~
		\subfigure[Occlusion and blurry faces without masks]{\label{fig:fig1_blurry}
			\includegraphics[width = 0.3\linewidth, height = 3cm]{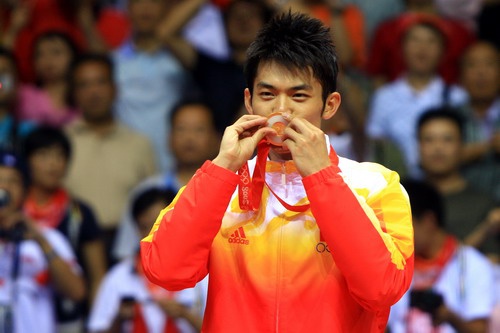}
		}
		~
		\subfigure[Makeup faces without masks]{\label{fig:fig1_makeup}
			\includegraphics[width = 0.3\linewidth, height = 3cm]{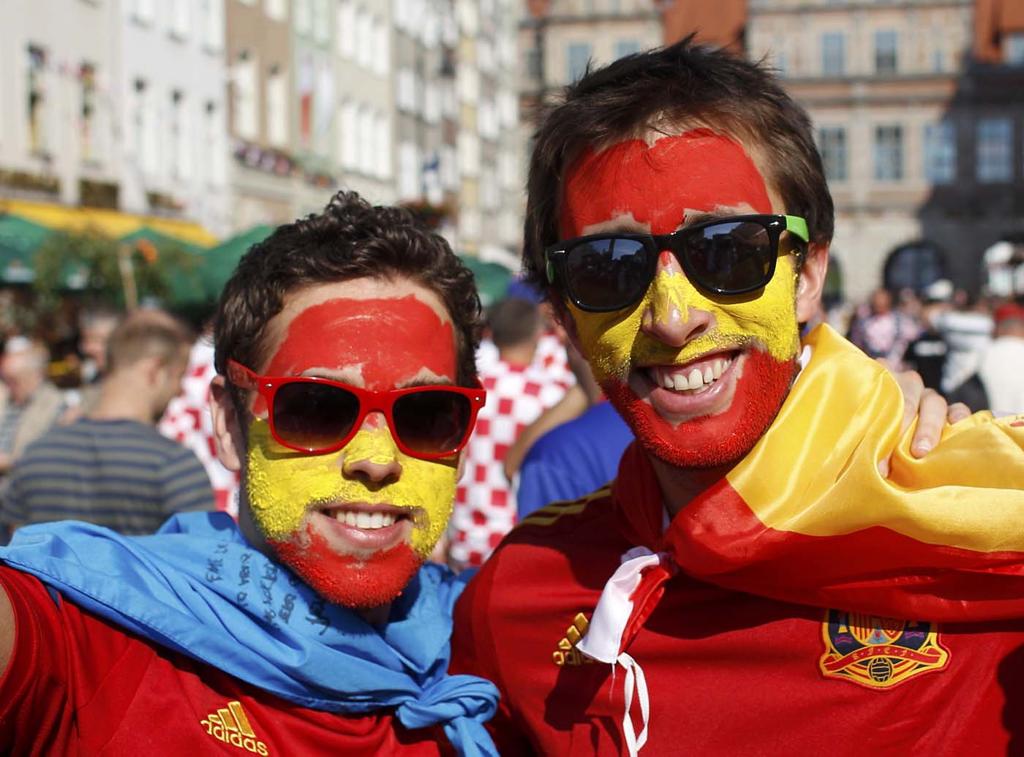}
		}
		~
		\subfigure[Masks with occlusion]{\label{fig:fig1_occlumask}
			\includegraphics[width = 0.3\linewidth, height = 3cm]{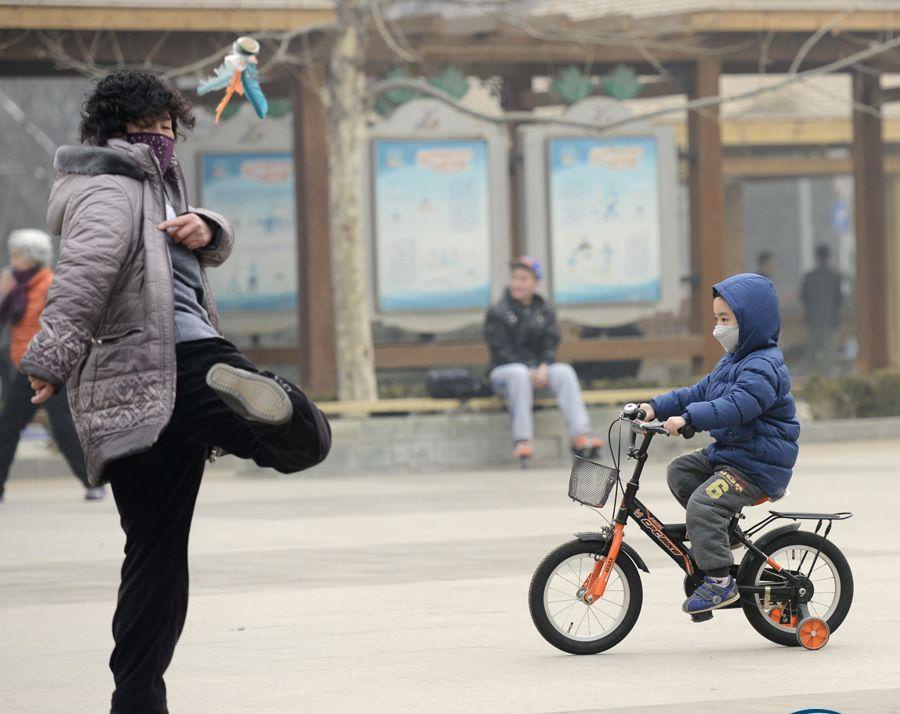}
		}
		\caption{Illustrative example of difficult cases on the AIZOO dataset.}
	\label{fig1} 
\end{figure}

With the advance of Artificial Intelligence (AI), it is promising to alleviate the burden by utilizing its machine learning algorithms for implementing an automatic monitoring system. Generally, facial mask detection can be regarded as a special case of object detection~\cite{zhao2019object,kumar2019face,jeong2019consistency, cao2020d2det, zheng2020distance, bochkovskiy2020yolov4, he2020momentum, tan2020efficientdet, kong2020foveabox, zhang2020bridging, yang2019reppoints, wang2020learning, chen2021you, mahmoud2020novel, lin2016masked, gathani2020detecting, chowdary2020face}. That is, the goal is to determine where the faces are (object localization) and whether the faces are with or without masks (object classification) in a given image. Object detection can be categorized into two mainstreams, anchor-based approaches~\cite{girshick2014rich,girshick2015fast,ren2016faster, he2020momentum, tan2020efficientdet, bochkovskiy2020yolov4, kong2020foveabox} and anchor-free approaches~\cite{redmon2016look,tian2019fcos,yang2019reppoints,chen2020reppoints, zhang2020bridging, lee2020centermask, wang2020learning}. Anchor-based approaches leverage predefined multiple sizes anchor boxes to detect objects with different scales and aspect ratios, which can also be categorized into one-stage detectors~\cite{redmon2016look,cao2020d2det, zheng2020distance, liu2016ssd, lin2017focal, fu2019retinamask} and two-stage detectors~\cite{girshick2015fast,ren2016faster, lin2017feature, cai2018cascade}. Two-stage detectors adopt region proposals to extract the region of interest (ROI) and separate the object detection task into object localization and image classification. Due to the high computational complexity of two-stage detectors, one-stage detectors merge the tasks of object localization and image classification into a regression problem by predicting class probabilities and bounding box coordinates simultaneously. On the other hand, anchor-free detectors truncate anchor boxes and directly detect the vital keypoints~\cite{redmon2016look,tian2019fcos,centernet, zhang2020bridging, yang2019reppoints, lee2020centermask, wang2020learning}, such as centers and corners of the object.


Generally, existing works of face detection solve a variety of challenging cases as illustrated in Figs.~\ref{fig:fig1_orientation}-\ref{fig:fig1_Crowded}, i.e., faces of different orientations, reflections, crowded scenes of different scales. Nevertheless, it is still challenging to detect faces with or without masks since several issues cannot be well-addressed by most of the previous works of object detection. First, masks are with different colors and styles, e.g., Fig.~\ref{fig:fig1_differentcolors} contains dark blue, light blue, white and pink masks, while the masks in Fig.~\ref{fig:fig1_poses} and Fig.~\ref{fig:fig1_nonhuman} are with different styles. To detect faces with masks of different colors and styles, one na\"{\i}ve solution is to increase the size of training data. However, the number of training data with facial masks is much smaller than of face images in existing datasets. Therefore, it is challenging to devise a machine learning model without a large-scale dataset. Second, images of wearing masks or not are only partially different, while the occlusions are similar to facial masks, e.g., one hand covering mouth in Fig.~\ref{fig:fig1_differentcolors}, kissing a medal in Fig.~\ref{fig:fig1_blurry}. Moreover, the occlusions of faces are diverse, e.g., makeup in Fig.~\ref{fig:fig1_makeup}, the clothing occluding the mask in Fig.~\ref{fig:fig1_occlumask}. One of the possible solutions is to use fine-grained feature extraction models, e.g.,~\cite{HRNET1, HRNET2, xie111, xie222}. Nevertheless, the embeddings of faces with and without facial masks, as well as occluded faces, are still relatively close to each other, which easily leads to the confusion of the model.

To address these issues, in this paper, we present a new framework, namely, "\verb|CenterFace|", for face mask detection with triplet-consistency representation learning. We first use the pre-trained models, e.g., ResNet-18~\cite{he2016deep}, MobileNet \cite{howard2017mobilenets}, to extract the basic features. 
Afterward, to enhance the basic features, the convolution block attention module \cite{woo2018cbam} is leveraged, which is the simple yet effective attention but can blend the cross-channel and spatial information together by operating the interactively informative features along the two significant dimensions. The input feature map can easily concatenate with the attention maps for adaptively refining the features in the feed-forward convolution neural networks module. As the anchor-free models demonstrate promising efficiency and effectiveness, the proposed "\verb|CenterFace|" uses a keypoint heat map to find the center point of its bounding box. Moreover, we propose a new Triplet-Consistency Representation Learning by integrating the consistency loss \cite{jeong2019consistency} and the triplet loss \cite{schroff2015facenet} to address the first and second challenges, respectively. Specifically, the consistency loss fully utilizes the labeled data to deal with the various orientations and visual appearance diversity, while the triplet loss emphasizes the difference of face with masks and occluded faces. Experimental results on public datasets~show that the proposed "\verb|CenterFace|" outperforms the state-of-the-art methods by 7.4\% and 6.3\% for the face with and without masks in terms of the precision, respectively, while the inference time satisfies the real-time constraint.

The main contributions can be summarized as shown bellows:
\begin{itemize}
    \item Since enforcing a mask policy is important to control the disease propagated through airborne transmission, e.g., COVID-19, and alleviates the burden of healthcare, we propose a new framework, namely, "\verb|CenterFace|", by utilizing the Triplet-Consistency Representation Learning to mitigate the effect of visual diversity and various orientations of the face masks.
    \item Experimental results on public datasets manifest that the proposed approach outperforms state-of-the-art methods, while the model size is small and can be adopted on the edge mobile devices. The source codes is released as a public download for public health improvement.
\end{itemize}
The rest of the paper is organized as follows. We present the related work in Section~\ref{sec:relatedwork}. The methodology is then described in detail in Section~\ref{sec:methodology}. Finally, the experimental results are elaborated in Section~\ref{sec:experiments} and we make conclusions and describe future work in Section~\ref{sec:conclusions}.


\section{Related works}
\label{sec:relatedwork}
\subsection{Object Detection}
Conventional face detection methods are based on the handcrafted feature due to the limitation of computing resources and the lack of large-scale datasets~\cite{viola2001rapid, viola2004robust,dalal2005histograms,lowe2004distinctive,lowe1999object,belongie2002shape,felzenszwalb2008discriminatively, felzenszwalb2009object,girshick2011object, girshick2012discriminatively, girshick2014rich, girshick2015fast}. For example, Viola et al.~\cite{viola2001rapid} propose the first real-time human face detector without any constraint by combining the Haar features selection with integral images and detection cascades. Dalal et al. \cite{dalal2005histograms} further propose an improved model by extracting the histogram of the oriented gradients (HOG)-based scale-invariant feature transformation, which are robust to the translation, scale and illumination features, as well as different sizes of objects \cite{lowe2004distinctive,lowe1999object,belongie2002shape}. However, to detect objects of different sizes, the HOG requires re-scaling input images several times. Therefore, Felzenszwalb et al. \cite{felzenszwalb2008discriminatively, felzenszwalb2009object} propose the extension of HOG with the advancement of Fourier Transform by integrating the Local Binary Pattern (LBP) to extract the scale-invariant features map. Girsick et al.~\cite{girshick2011object, girshick2012discriminatively, girshick2014rich, girshick2015fast} propose the notion that objects can be modeled by parts in a deformable configuration and ensemble the detection of different object parts for the final prediction. Nevertheless, the representative DFM consists of a root-filter and a number of part-filters instead of specifying the size and location of the part filters.

With the advance of deep learning, extracting features from images is now data-driven instead of using the prior knowledge to handcraft the features. Object detection can be grouped into two categories: one-stage detector and two-stage detector. The two-stage detectors \cite{girshick2014rich, ren2015faster, girshick2015fast, sun2015deeply, taigman2014deepface, zhu2014recover, joshi2020deep} first extract a set of object candidate boxes by selective search algorithm \cite{van2011segmentation} as region proposals, and region proposals are warped into a square and fed into a convolutional neural network for extracting discriminative features. However, a large number of overlapped region proposals make redundant feature computation, leading to inefficient detection models. Therefore, several works make progress on improving the computing of the overlapped proposals \cite{ren2015faster, girshick2015fast,lin2017feature,ren2015faster}. Even though, the training process is still multi-stage with computation redundancy at the subsequent detection stage.

On the other hand, one-stage detectors can be further categorized into anchor-based methods and anchor-free methods. Redmon et al. propose you only look once (YOLO) series~\cite{redmon2016you, redmon2017yolo9000, redmon2018yolov3,yolov4}, which replace the region proposal network (RPN) with anchor boxes predefined the ratio of width and heights of objects. However, YOLO series still suffers from improving the localization accuracy as compared with two-stage detectors. To solve the problem of one stage detection on localization and small object detection, Liu et al. propose a single shot multibox detector (SSD) \cite{liu2016ssd}, which focuses on the multi-scale object detection with a variety of sizes and aspect ratios and computes both the location and class scores using small convolution filters. Despite the improvement of the speed and accuracy, the performance of one-stage detectors is usually inferior to two-stage detectors. A recent line of studies proposes anchor-free detectors, which directly find the objects without using multiple anchors in the input images~\cite{lin2017focal,law2018cornernet, zhou2019bottom,kong2020foveabox, tian2019fcos}. There are two popular kinds of anchor-free detection methods. The first kind is keypoint-based methods \cite{law2018cornernet, zhou2019bottom}, which bounds the spatial extent of objects by locating several predefined or self-learned keypoints. The second one is center-based methods \cite{kong2020foveabox, tian2019fcos}, which defines positive by the center point or region of objects to predict their boundary.

{ Self-supervised learning has become a popular learning, which creates pretext tasks on unlabeled data and learns in a supervised manner. In other words, the goal of self-supervised learning is to construct the representation of images with semantically meaningful content via pretext task while not requiring the semantic annotation for a large training set of images. For example, Misra et al.~\cite{selfsupervisedQ} use the pretext tasks with image transformation to encourage the representation of images to be invariant with the image patch perturbation. Xu et al.~\cite{ssECCV2020} exploit the similarity from the self-supervised signals as an auxiliary task, which can effectively transfers the hidden information from the teacher to the student network. Gidaris et al.~\cite{selfsupervisedS} propose a self-supervised learning method based on transforming input images in different rotations, so the framework can learn to estimate the geometric transformations applied to the image, which helps the downstream tasks, such as object detection. Grill et al.~\cite{selfsupervisedT} leverage two networks, where an online network is trained to predict the representation of another network of another augmented view of the identical image.}

\subsection{Attention Mechanism}
Attention has become a popular concept and a useful tool in the deep learning community in recent years \cite{vaswani2017attention, wang2017residual,hu2018squeeze,woo2018cbam, elsayed2019saccader, dai2019second, locatello2020object}. The basic idea is that human visual perception only focuses on some specific regions at one time and performs well in object detection. Therefore, to mimic human attention as a sequence of partial glimpses, different kinds of attention mechanisms are proposed to learn ``what'' and ``where'' to attend and then focus on the important features and suppressing unnecessary ones. For example, Vaswani et al. \cite{vaswani2017attention} use a decoder-encoder architecture with the attention mechanism to refine the feature maps. Hu et al. \cite{hu2018squeeze} introduce the inter-channel attention by using global-pooled features to compute their channel-wise attention. Woo et al. \cite{woo2018cbam} demonstrate that channel-wise attention is insufficient and provides the spatial attention to decide ``where'' to focus, enabling the attention generation process for 3D features map with much less number of parameters.

\begin{figure}[t] 
  \centering
  \includegraphics[width=\linewidth]{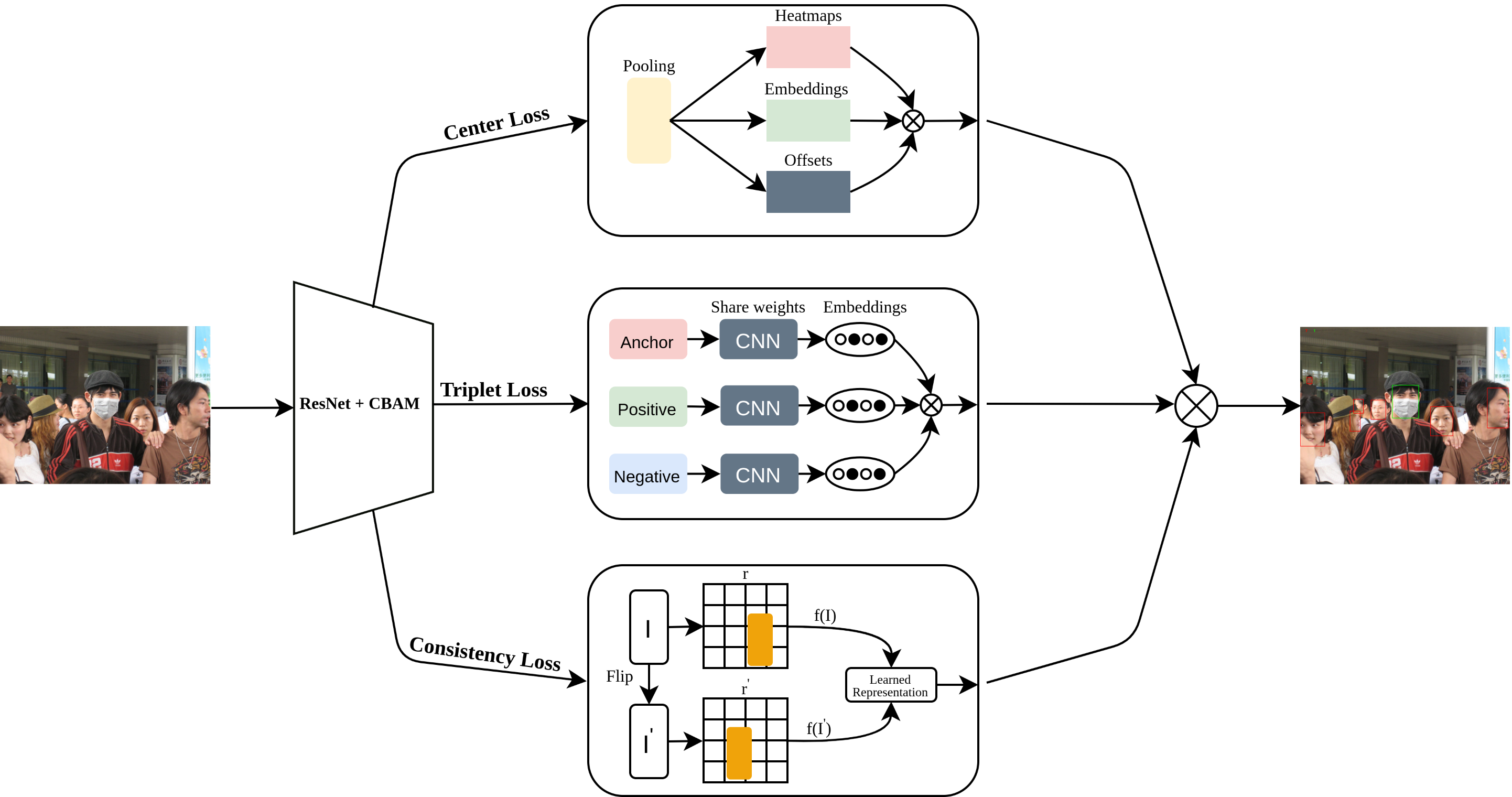} 
  \caption{The overview architecture of CenterFace. We use an output heatmap from a convolution network to detect a face and/or a mask as a pair of the bounding box, the network is trained to predict similar embeddings for faces that belong to the same face. }
  \Description{Overview of the proposed CenterFace framework.}
  \label{fig4}
\end{figure}
\section{Methodology}\label{sec:methodology}

\begin{figure}[t] 
  \centering
  \includegraphics[width=\linewidth]{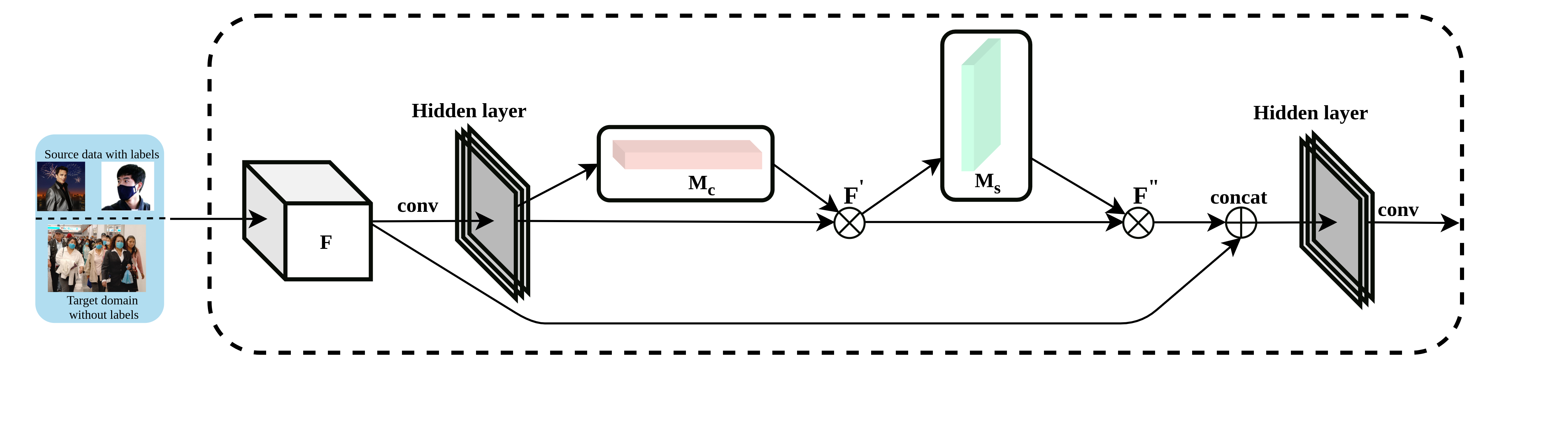}
  \caption{Convolution Block Attention Module (CBAM) \cite{woo2018cbam} with a block in ResNet~\cite{he2016deep}. We utilize CBAM to integrate within the convolution outputs in each ResNet block.}
  \label{fig3}
\end{figure}

To effectively detect the faces with and without facial masks, we present a new framework, namely, "\verb|CenterFace|", for face mask detection with triplet-consistency representation learning. Fig.~\ref{fig4} shows the overview of the proposed "\verb|CenterFace|". Specifically, we first use the pre-trained models to extract the low-level features since low-level feature extraction is usually shared in different tasks, e.g., texture, edges. Afterward, to enhance the basic features without significantly increasing the number of parameters, the convolution block attention module~(CBAM)~\cite{woo2018cbam} is leveraged to attend the channel and spatial features. Fig.~\ref{fig3} illustrates the model architecture of CBAM, which contains channel and spatial modules. The channel module utilizes both max-pooling outputs and average-pooling outputs for finding the attention weights on channels, while the spatial module uses similar two outputs and pool along the channel axis to generate the attention weights on spatial dimensions.

Moreover, since the anchor-free models have been proved to be efficient and effective, the proposed "\verb|CenterFace|" uses a key-point heat map to find the center point of its bounding box. Moreover, we propose a new Triplet-Consistency Representation Learning by integrating the consistency loss \cite{jeong2019consistency} and the triplet loss \cite{schroff2015facenet} to address the first and second challenges, respectively. Specifically, the consistency loss fully utilizes the labeled data to deal with the various orientations and visual appearance diversity, while the triplet loss emphasizes the difference of face with masks and occluded faces. In the following, we first explain the center loss, which leverages the anchor-free models, and then the triplet loss and consistency loss for Triplet-Consistency Representation Learning. Finally, the total loss function is presented, together with the pseudocode. { In the following, bold uppercase letters (e.g., $\mathbf{X}$) and lowercase letters (e.g., $\mathbf{x}$) denote matrices and column vectors, respectively. Non-bold letters (e.g., $x$) and squiggle letters (e.g., $\mathcal{X}$) represent scalars and tensors, respectively.}

\subsection{Center Loss}
Most of the successful object detection models require enumerating candidate object locations and classifying each candidate region, which is computationally expensive. Moreover, since the model may predict several candidate bounding boxes for one object, the predicted bounding boxes require additional post-processing, e.g., non-maximum suppression~\cite{nms2006}. To satisfy the requirement of real-time monitoring, we adopt the anchor-free method, which performs object classification and bounding box localization simultaneously. { Specifically, let $\mathcal{I} \in \mathbb{R}^{h{\times}w{\times}3}$ denote the input image tensor with the width $w$ and height $h$. The goal is to generate the key-point heatmap $\hat{\mathcal{Y}}\in  [0,1]^{\frac{h}{s}{\times}\frac{w}{s}{\times}c}$ with the stride $s$, where $\hat{\mathcal{Y}}_{i,j,k} = 1$ and $\hat{\mathcal{Y}}_{i,j,k} = 0$ respectively represent a predicted keypoint and the background region at position $(i,j)$ belonging to class $k$, $c$ is the number of keypoint classes,\footnote{We set $s$ as $4$ according to previous work \cite{cao2017realtime, newell2016stacked, papandreou2017towards}, and set $c$ as $2$ since the goal is to detect the faces are with or without masks.}}

To train the model for predicting the keypoints, given a ground truth keypoint $\textbf{p}=(p_x,p_y)\in \mathbb{R}^{2}$, we first calculate the low resolution equivalence $\Tilde{\mathbf{p}}=(\lfloor\frac{p_x}{s}\rfloor,\lfloor\frac{p_y}{s}\rfloor$). We then smooth the groundtruth keypoint by the 2D Gaussian kernel to derive the heatmap $\mathcal{Y}\in[0,1]^{\frac{h}{s}{\times}\frac{w}{s}{\times}c}$ as follows.

\begin{equation} \label{firstlossa:initilize}
  y_{i,j,k} = \exp\left ( - \frac{(i-\left \lfloor  \frac{p_{x}}{s}\right \rfloor)^{2} + (j-\left \lfloor \frac{p_{y}}{s} \right \rfloor)^{2}}{2\delta_{p}^{2}}\right ),
\end{equation}
where $\delta_{p}$ is the object-size standard deviation, which is determined by its size to ensure that the pair of radius points can generate the bounding box around the face by the ground truth annotation for each ground truth keypoint $\mathbf{p}$.

The training objective is a pixelwise logistic regression with a variant of focal loss \cite{lin2017focal}, i.e.,
\begin{equation} \label{firstlossa}
L_{pix} = -\frac{1}{N}\sum_{i=1}^{h}\sum_{j=1}^{w}\sum_{k=1}^{c}\left\{\begin{array}{ll}
(1-\hat{Y}_{i,j,k})^{\alpha}\log(\hat{Y}_{i,j,k}) & \text{if $Y_{i,j,k}$=1,} \\ 
(1-Y_{i,j,k})^{\beta}(\hat{Y}_{i,j,k})^{\alpha}\log(1-\hat{Y}_{i,j,k}) & \text{otherwise,}
\end{array}\right.
\end{equation}
where $\alpha$ and $\beta$ are the hyperparameters controlling the contribution for each point of the focal loss.\footnote{Empirically, we set $\alpha$ and $\beta$ as $2$ and $4$, respectively.} $N$ is the number of the faces in input images $I$, which is applied to normalize all the positive focal instance losses to 1.

{ As the output stride may cause errors due to the discretization, i.e., $\mathbf{p}$ becomes $\mathbf{\Tilde{p}}$, we predict the local offset $\hat{\mathcal{O}}\in \mathbb{R}^{\frac{h}{s}\times\frac{w}{s}\times2}$ for each center point, where the groundtruth offset can be derived by calculating $\frac{\mathbf{p}}{s}-\lfloor\frac{\mathbf{p}}{s}\rfloor$ for each center point $\mathbf{p}$. Let $\mathbf{p}_n$ denote the position of the $n$-th center. The offset loss can be derived as follows.

\begin{equation} \label{firstlossc}
    L_{off} = \frac{1}{N}\sum_{n=1}^{N}\left | \hat{\mathcal{O}}_{\mathbf{p}_n} - (\frac{\mathbf{p}_n}{s}-\lfloor\frac{\mathbf{p}_n}{s}\rfloor)\right |.
\end{equation}

In addition to the centers derived by the keypoint heatmap $\hat{\mathcal{Y}}$ and offset $\hat{\mathcal{O}}$, the next goal is to find the bounding boxes. Let $(x_{1}^{n}, y_{1}^{n}, x_{2}^{n}, y_{2}^{n})$ be the coordinates of the bounding box of the object $n$ with class $c_{n}$, while the center can be represented by $\mathbf{p}_{n} = \left ( \frac{x_{1}^{n}+x_{2}^{n}}{2}, \frac{y_{1}^{n}+y_{2}^{n}}{2} \right )$. We predict the bounding box size $\hat{\mathcal{S}}_{n}\in \mathbb{R}^{\frac{h}{s}\times\frac{w}{s}\times2}$ for each center point. It is worth noting that $\hat{\mathcal{S}}_{n}$ is shared for the classes of the faces with or without masks to reduce the computational cost. The groundtruth for the $n$-th center can be calculated by $\mathbf{s}_{n}=(x_{2}^{n} - x_{1}^{n}, y_{2}^{n} - y_{1}^{n})$. As such, the loss for bounding box size, denoted by $L_s$, can be derived as follows.
\begin{equation} \label{firstlossb}
    L_{s} = \frac{1}{N}\sum_{n=1}^{N}\left | {\hat{S}_{p_{n}}}-\mathbf{s}_{n} \right |.
\end{equation}
The overall center loss for detecting faces with and without masks is: 
\begin{equation} \label{1stloss}
    L_{center} = \lambda_{pix} L_{pix} + \lambda_{off} L_{off}+\lambda_{s}L_{s},
\end{equation}
where $\lambda_{pix}$, $\lambda_{s}$, and $\lambda_{off}$ are the hyperparameters controlling the weights between different prediction targets.\footnote{The hyperparameters are empirically set as
$\lambda_{pix} = 1$, $\lambda_{off} = 1$ and $\lambda_{s} = 0.01$.}}

\subsection{Triplet Loss}
Due to the subtle difference between faces with masks and faces with occlusions, it is difficult to train a model by only using the center loss. Therefore, to further improve the performance on challenging cases, we utilize an online triplet mining method \cite{schroff2015facenet}, which enforces that the distance between a pair of samples with the same label is smaller than that between a pair of samples with different labels. Indeed, the triplet loss works directly on features. A triplet is formed by 1) anchor input, 2) positive input (samples with the same label) and 3) negative input (samples with the different labels or random cropped samples). The distance between the anchor input and the positive input is enforced to be smaller than the distance between the anchor input and the negative input. {Specifically, let $\mathcal{R}^a_i$, $\mathcal{R}^{+}_i$ and $\mathcal{R}^{-}_i$ respectively denote anchor image region, positive image region and negative image region in the $i$-th triplet. The triplet loss can be derived as follows.
\begin{equation} \label{2ndloss} 
    L_{tri} = \sum_{i}\left [ \left \| f(\mathcal{R}_{i}^{a}) - f(\mathcal{R}_{i}^{+}) \right \|_{2}^{2} - \left \| f(\mathcal{R}_{i}^{a}) - f(\mathcal{R}_{i}^{-}) \right \|_{2}^{2} + \epsilon \right ].
\end{equation}
The input image regions $\mathcal{R}_{i}^{a}$, $\mathcal{R}_{i}^{+}$ and $\mathcal{R}_{i}^{-}$ are embedded by the same mapping function $f$. We ensure that the features of the anchor region $\mathcal{R}_{i}^{a}$ is particularly close to other positive samples $\mathcal{R}_{i}^{+}$ and far away from any of the negative images 
$\mathcal{R}_{i}^{-}$ by using the margin $\epsilon$ to separate positive and negative pairs. In other words, when faces are regarded as anchors and positive examples, the masks and some regions near the face and mask areas become negative examples. On the other hand, when the face with masks are regarded as anchors and positive examples, the face area and some regions near face areas become negative examples. Eventually, the distance from the anchor input to the positive samples is minimized, while the distance from the anchor input to the negative samples is maximized.}

\subsection{Consistency Loss}
In addition to the triplet loss, we further impose the consistency constraint on the predicted heatmap of the original image and the predicted heatmap of the horizontally-flipped images to simultaneously stabilize the prediction results and enlarges the training data. The consistency constraint is especially effective when incorporating the triplet loss since multiple positive samples can be created at the same time. Specifically, let $\hat{\mathcal{Y}}$ and $\hat{\mathcal{Y}}'$ denote the predicted heatmap of the original image and the predicted heatmap of the horizontally-flipped images, respectively. The prediction probability at position $(i,j)$, i.e., $\hat{\mathcal{Y}}_{i,j,:}$, should be close to the prediction probability at the horizontally opposite position $(i',j')$, i.e., $\hat{\mathcal{Y}}'_{i',j',:}$. One way to measure the closeness is to use L2 distance. However, L2 loss regards all the classes equally so that the irrelevant classes with a low probability may also highly affects the results. {Therefore, inspired by \cite{jeong2019consistency}, we also use Jensen-Shannon divergence (JSD) to measure the difference between $\hat{\mathcal{Y}}_{i,j,:}$ and $\hat{\mathcal{Y}}'_{i',j',:}$. The consistency loss for classification, denoted by $L_{con-c}$, is obtained by calculating the expectation of all bounding box pairs can thus be derived as follows.
\begin{equation} \label{3rdeq1}
    L_{con-c} = \mathbb{E}_{n}[JSD(\hat{\mathcal{Y}}_{\mathbf{p}_n},\hat{\mathcal{Y}}'_{\mathbf{p}^\prime_n}))].
\end{equation}
On the other hand, the localization result of the candidate box is based on the offset $\hat{\mathcal{O}}$ and size $\hat{\mathcal{S}}$. Let $\hat{\mathcal{O}}'$ and $\hat{\mathcal{S}}'$ denote the predicted offset and size of the horizontally-flipped images, respectively. Unlike the classification, a simple modification is required to make the prediction equivalent to each other. As the flipping transformation makes the offset change in the opposite direction, a negation is applied to correct the groundtruth, i.e., 
\begin{equation*}
    \hat{\mathcal{O}}_{\mathbf{p}_n,1} \Leftrightarrow -\hat{\mathcal{O}}'_{\mathbf{p}'_n,1} 
\end{equation*}
\begin{equation*}
    \hat{\mathcal{O}}_{\mathbf{p}_n,2}, \hat{\mathcal{S}}_{\mathbf{p}_n} \Leftrightarrow \hat{\mathcal{O}}'_{\mathbf{p}'_n,2}, \hat{\mathcal{S}}'_{\mathbf{p}'_n}
\end{equation*}  
Afterward, we derive the localization consistency loss by calculating the expectation of the offset and size difference for each single pair of the candidate box center at position $\mathbf{p}_n$ and $\mathbf{p}'_n$ as below.
\begin{equation}\label{3rdeq2}
    L_{con-l} =\mathbb{E}_{n}[\left \| \hat{\mathcal{O}}_{\mathbf{p}_n,1}-(- \hat{\mathcal{O}}'_{\mathbf{p}'_n,1}) \right \|^{2}+\left \| \hat{\mathcal{O}}_{\mathbf{p}_n,2}-( \hat{\mathcal{O}}'_{\mathbf{p}'_n,2}) \right \|^{2}+\left \| \hat{\mathcal{S}}_{\mathbf{p}_n}- \hat{\mathcal{S}}'_{\mathbf{p}'_n} \right \|^{2}].
\end{equation}
}
Finally, when the consistency loss is computed with all candidates, the results may easily be dominated by the backgrounds, which deteriorates the performance of the foreground classification candidates. As such, we have to build a mask with the same size as the heatmap for every groundtruth bounding box to exclude the boxes with a high probability of background class. The total consistency loss, denoted by $L_{con}$, simply summarizes $L_{con-c}$ and $L_{con-l}$, i.e., $L_{con} = L_{con-c} + L_{con-l}$.

\subsection{Overall Objective Function}
The total loss function is summarized over the center loss, triplet loss and consistency loss, i.e.,
\begin{equation} \label{totalloss}
    L_{total} = L_{center}  + \lambda_{tri} L_{tri} + \lambda_{con} L_{con},
\end{equation}
where $\lambda_{tri}$ and $\lambda_{con}$ are hyper-parameters controlling the contribution of different loss terms.\footnote{Empirically, we set $\lambda_{tri} = 1$ and $\lambda_{con} = 100$.} In summary, the keypoint heatmap prediction works as a general-purpose object detector which extends the keypoint estimator to generate the face's bounding box. On the other hand, the triplet loss, which enforces the face margin between each pair of face to roughly align matching/non-matching face, along with the consistency classification and localization loss are proposed to localize not only the face classification but also the position. The overall objective loss facilitates the detection model to be robust enough for any complicated situations among various representation of the face detection, especially from the masked face detection. The pseudocode of "\verb|CenterFace|" is presented in Algorithm \ref{pseudo:alg1}.

{
\begin{algorithm}
\caption{CenterFace algorithm}
\begin{algorithmic}
\State $\textbf{Input:}$
\State \tab Input image set with groundtruth 
\State \tab Hyperparameters $\alpha$ and $\beta$ (focal loss), stride $s$, margin $\epsilon$, $\lambda_{pix}$, $\lambda_{s}$, $\lambda_{off}$, $\lambda_{tri}$, $\lambda_{con}$, number of iterations $n_{iter}$
\State $ \textbf{Output:}$ 
\State \tab Bounding box with class represented by $\hat{\mathcal{Y}}^\ast$, $\hat{\mathcal{S}}^\ast$ and $\hat{\mathcal{O}}^\ast$
\end{algorithmic}
\begin{algorithmic}[1]


\For{$\textit{iter}$ = $1$...$n_{iter}$}
\State Generate heatmaps $\hat{\mathcal{Y}}$ with size $\hat{\mathcal{S}}$ and offset $\hat{\mathcal{O}}$ from $L_{center}$ according to Section 3.1
\For{$\textbf{each}$ $\mathcal{I}$}
\State \tab Compute $L_{tri}$ according to Eq.~\ref{2ndloss} 
\State \tab Compute $L_{con-c}$ according to Eq.~\ref{3rdeq1} 
\State \tab Compute $L_{con-l}$ according to Eq.~\ref{3rdeq2}
    \State \tab Optimize $L_{total} = L_{center}  + \lambda_{tri} L_{tri} + \lambda_{con} L_{con}$
\EndFor
\If{$L_{total}$ is the smallest so far}
    \State $\hat{\mathcal{Y}}^\ast$, $\hat{\mathcal{S}}^\ast$, $\hat{\mathcal{O}}^\ast$ $\leftarrow$ $\hat{\mathcal{Y}}$, $\hat{\mathcal{S}}$, $\hat{\mathcal{O}}$
    \EndIf
\EndFor
\State $\textbf{return}$ $\hat{\mathcal{Y}}^\ast$, $\hat{\mathcal{S}}^\ast$ and $\hat{\mathcal{O}}^\ast$

\end{algorithmic}
\label{pseudo:alg1}
\end{algorithm}
}

\section{Experiments}\label{sec:experiments}
In this section, we first present the experimental setup in detail. Next, we evaluate the proposed framework on public datasets with diverse face mask representations including synthetic data, visual diversity, and various orientations. Finally, the ablation studies are presented to show the effectiveness of each component in the proposed framework.

\subsection{Experimental Setup}
\noindent\textbf{Datasets.} In this section, we evaluate our detection approach on the AIZOO dataset \cite{aizootech}, WiderFace \cite{yang2016wider} and MAFA dataset \cite{ge2017detecting} to verify the effectiveness of our "\verb|CenterFace|" module with baselines. AIZOO dataset is introduced when the pandemic situation just ransomed for the hope that people in the world can defeat the pandemic as soon as possible. The data set includes 7,959 images with various illuminations, occlusion, and different poses, and splits into the training, testing, and validation sets with 4096, 1,226, and 1,839, respectively. Moreover, WiderFace dataset contains 393,703 annotated faces with a variety in poses, crowd, face expression, occlusion, and scenarios from 32,203 images. These images are split into three subsets: training (40\%), validation (10\%), and testing (50\%) set along with three difficulty levels: easy, medium, and hard based on the detection rate of EdgeBox benchmark \cite{zitnick2014edge}. On the other hand, MAFA Dataset is a popular facial mask dataset that contains 35,806 masked faces in 30,811 images. This image dataset has been covered almost incident cases of occluded faces in real-life. In detail, it contains 60 cases of occluded faces with three different levels of occlusions in daily scenarios. Furthermore, four types of masks and five face orientations are also provided to diversify the dataset, and if the faces are considerably blurred or smaller than 32 pixels, it is ignored. All of those properties are divided into three subsets: the whole subset contains 6354 masked faces, 996 unmasked faces, and 2683 ignored faces. The masked subset and the remaining subset consist of both masked faces and unmasked faces.

\noindent\textbf{Baselines.} To the best of our knowledge, only one published paper that focuses on face mask detection with available source codes. Specifically, in AIZOO dataset, the authors verify their dataset by deploying the structure of SSD \cite{liu2016ssd} with a light-weight backbone network, which contains 24 layers, of which 8 layers are convolution layers. The input size is set as $260 \times 260$, while the total number of parameters is only 1.01M.

\noindent\textbf{Evaluation Metrics}
We evaluate the above approach by calculating the precision and recall metrics for the facial detection model and masked facial classifier, which is defined as follows.
\begin{equation}
\begin{matrix}
Precision = \frac{TP}{TP+FP}\times 100\% \\[0.5cm]
Recall = \frac{TP}{TP+FN}\times 100\% \\[0.5cm]
\end{matrix} \label{matrix11}
\end{equation}

\begin{table}
    \caption{Detection Accuracy of CenterFace}
    \label{tab:commands}
    \centering\sffamily
        \begin{tabular}{ |c|c|c|c|c| }
        \hline
        \multirow{2}{*}{\bf{Model}} &\multicolumn{2}{c|}{\bf{Face}} &\multicolumn{2}{c|}{\bf{Mask}} \\ \cline{2-5}
        &\multicolumn{1}{c|}{Precision} &\multicolumn{1}{c|}{Recall} 
        &\multicolumn{1}{c|}{Precision} &\multicolumn{1}{c|}{Recall}  \\
        \hline
        baseline \cite{aizootech}&   89.6\%    & 85.3\%   &91.9\%  &88.6\% \\
        RetinaMask + MobileNet \cite{jiang2020retinamask}     &83.0\%  & 95.6\%   &82.3\% &89.1\%  \\
        RetinaMask + ResNet50 \cite{jiang2020retinamask} &91.9\% & \textbf{96.3\%} &  93.4\% & 94.5\%\\
        CenterFace (MobileNet) &   96.8\%  & 88.0\% & 88.0\% &95.2\% \\
        CenterFace (ResNet18)    &\textbf{99.3\%} & 96.2\%&  \textbf{99.7\%} & \textbf{96.7\%} \\
        \hline
        \end{tabular}
        \label{table:accuracy}
\end{table}

\subsection{Implementation Details}
In the experiment, we follow the evaluation protocol in \cite{xiao2018simple}, and use ResNet-18 \cite{he2016deep} as the backbone network. The model is trained on an input image resolution $520 \times 520$ with a batch size of 10 with Adam optimizer \cite{kingma2014adam}. The learning rate is $1.25e^{-4}$ for 140 epochs, with the learning rate dropped $10\times$ at 30 and 80 epochs, respectively. We implement the proposed CenterFace with Pytorch \cite{paszke2019pytorch}. Then, to strengthen the effective attention for feed-forward convolutional neural network, we utilize CBAM \cite{woo2018cbam} to every single block of the CNNs. The attention module not only improves the effectiveness of the CNNs but also gives a solution to reduce the computation and time consuming of the framework.

\subsection{Experiment Results}
\noindent\textbf{Performance Comparison on AIZOO Dataset.} We evaluate the proposed model with two light-weight backbone networks, ResNet-18 \cite{he2016deep} and MobileNet \cite{howard2017mobilenets}, which are capable to deploy on the edge mobile devices. Table \ref{table:accuracy} compares the proposed method with different approaches in terms of precision and recall. The results manifest that our model significantly outperforms other candidate methods. Specifically, equipped with the proposed Triplet-Consistency Representation Learning, MobileNet and ResNet improve the RetinaMask \cite{jiang2020retinamask} by 7.4\% and 6.3\% in both face and mask classes in terms of precision. On the other hand, the recall of the proposed approach is almost equaled to the face class and 2.2\% higher in the mask class.

\begin{table}
    \caption{Flops and Number of Parameters of Models (AIZOO dataset).}
    \label{tab:commands}
    \centering\sffamily
        \begin{tabular}{ |c|c|c|}
        \hline
        \bf{Model} &\bf{FLOPs} &\bf{MACs} \\\hline
        baseline \cite{aizootech} &1.01 M &--\\
        RetinaMask + MobileNet \cite{jiang2020retinamask} &3.5 M &1.67 G\\
        RetinaMask + ResNet50 \cite{jiang2020retinamask} &25.56 M &21.53 G\\
        CenterFace (MobileNet) &3.5 M &1.67 G \\
        CenterFace (ResNet18) &11.69 M &9.52 G \\
        \hline
        \end{tabular}
        \label{table:flop}
\end{table}

\noindent\textbf{Efficiency Analysis.} To evaluate the efficiency of different approaches, we use FLOPs and MACs to measure the complexity of different models. As shown in Table~\ref{table:flop}, the results manifest that the number of parameters for ours with RestNet-18 network is much less than the state-of-the-art results using ResNet-50 since the model size of ResNet-50 is greater than the of ResNet-18. Nevertheless, even with a deep network, Table \ref{table:state} still shows that a shallower network with the proposed triplet-consistency representation learning performs better than that with a deeper network. Even more, when using the same MobileNet network, our result still outperforms the current state-of-the-artwork. This is because of the efficiency of the anchor-free techniques and the attention module we utilize in this work, which saves the computation cost by enhancing the features with fewer parameters, instead of directly increasing the number of layers.

\begin{figure}
    \centering
    \subfigure[Crowd scene with different scales]{\label{fig:fig5_crowd}
    \includegraphics[width = 0.2\linewidth, height = 2cm]{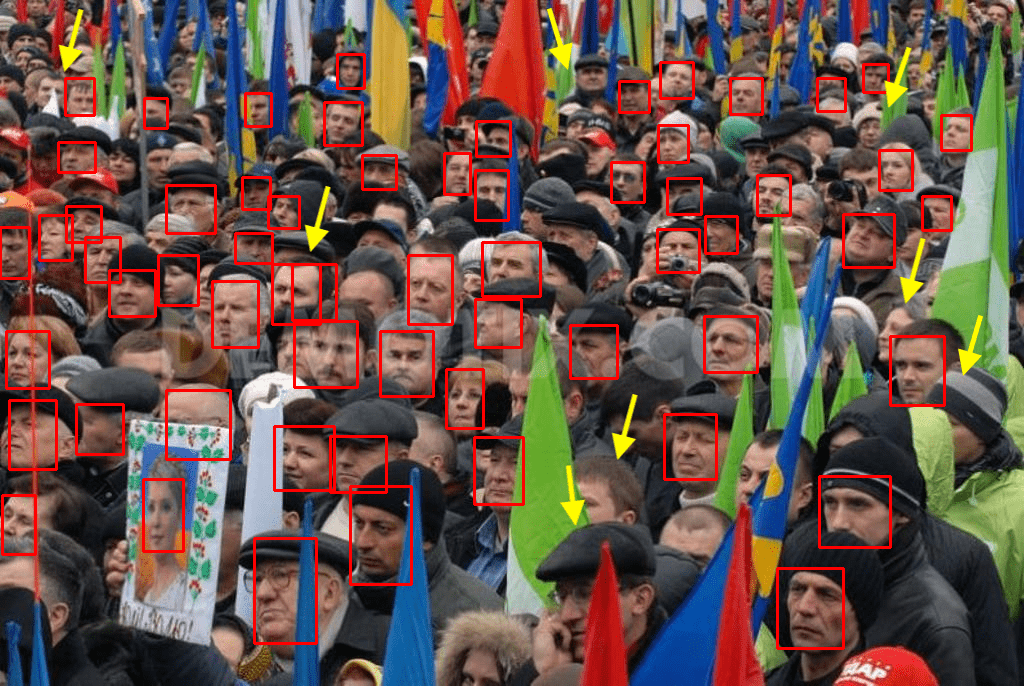}
    ~
    \includegraphics[width = 0.2\linewidth, height=2cm]{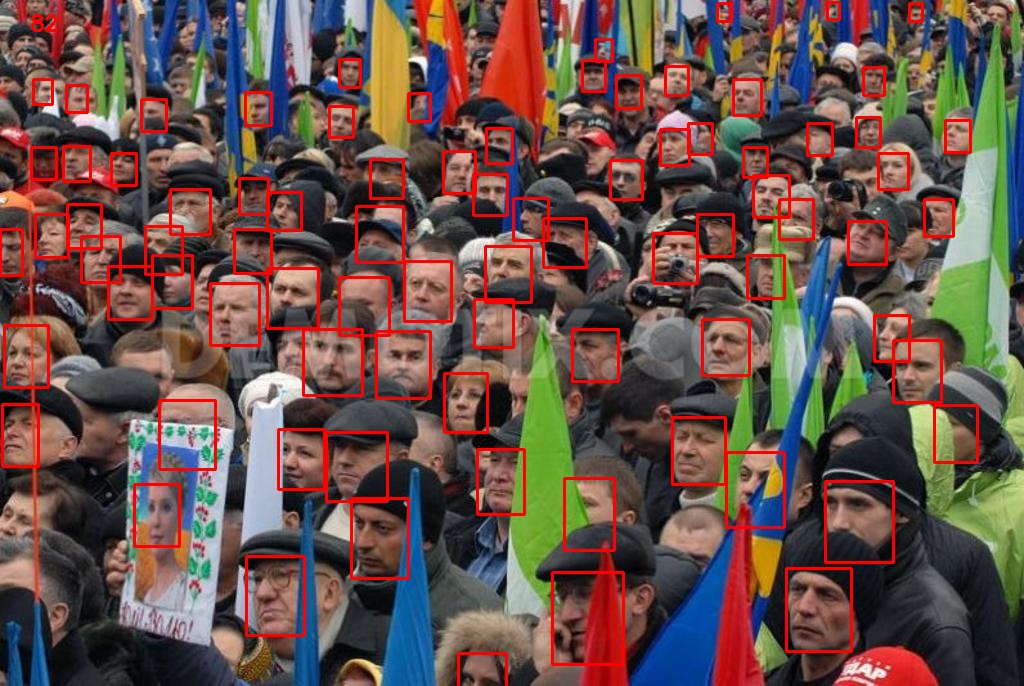}}
    \subfigure[Multi-pose facial detection]{\label{fig:fig5_poses}
    \includegraphics[width = 0.2\linewidth, height = 2cm]{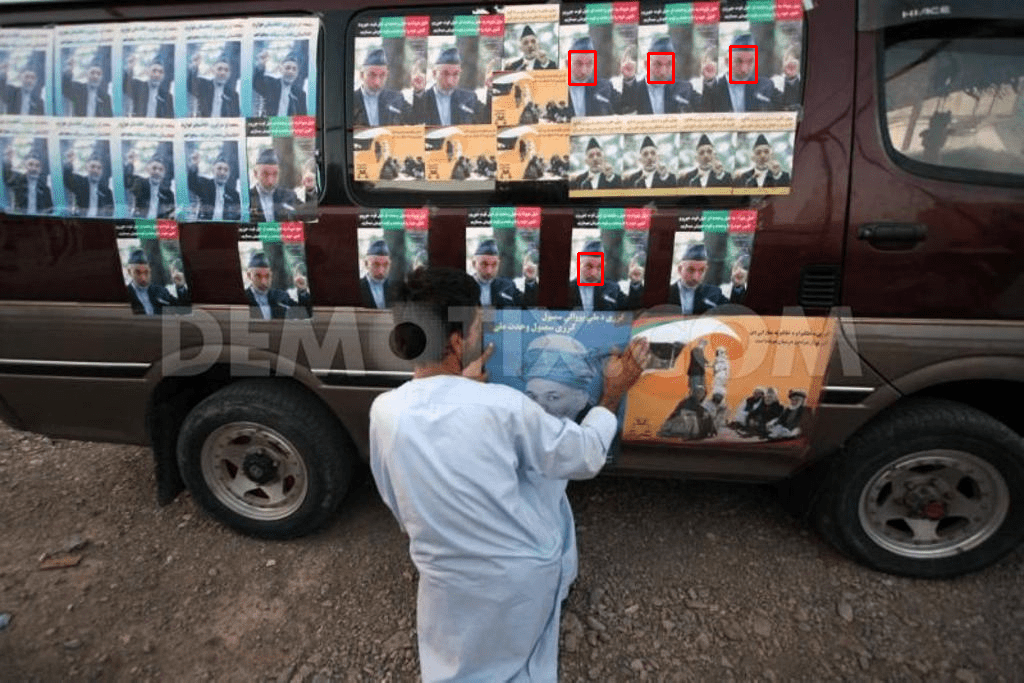} ~
    ~
    \includegraphics[width=0.2\linewidth, height=2cm]{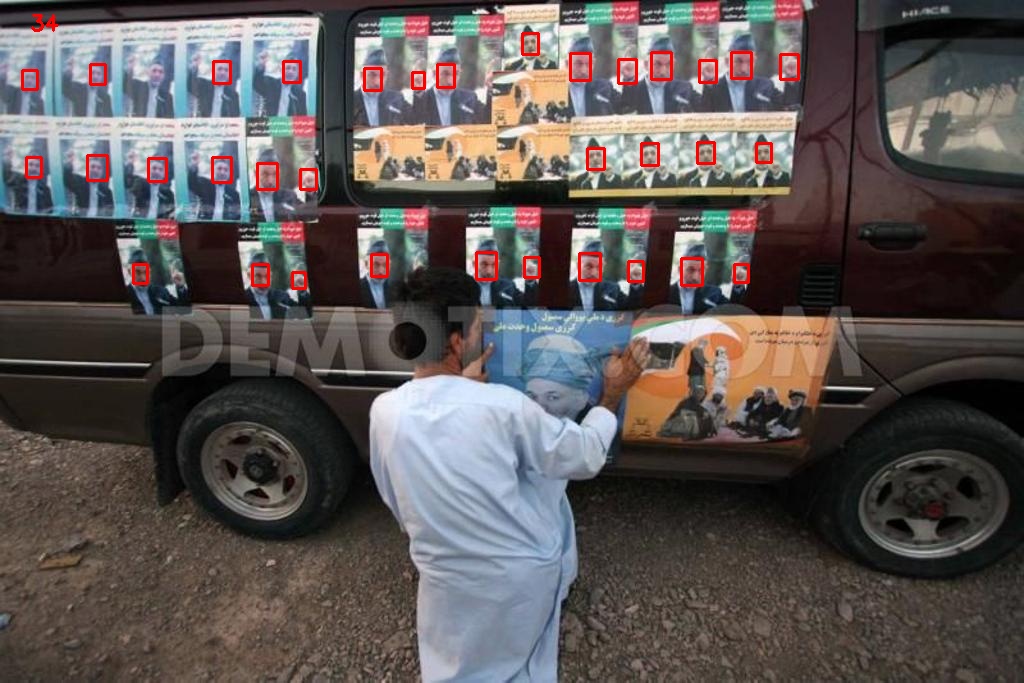}} 
    \subfigure[Occlusion faces]{\label{fig:fig5_hiddens}
    \includegraphics[width = 0.2\linewidth, height = 2cm]{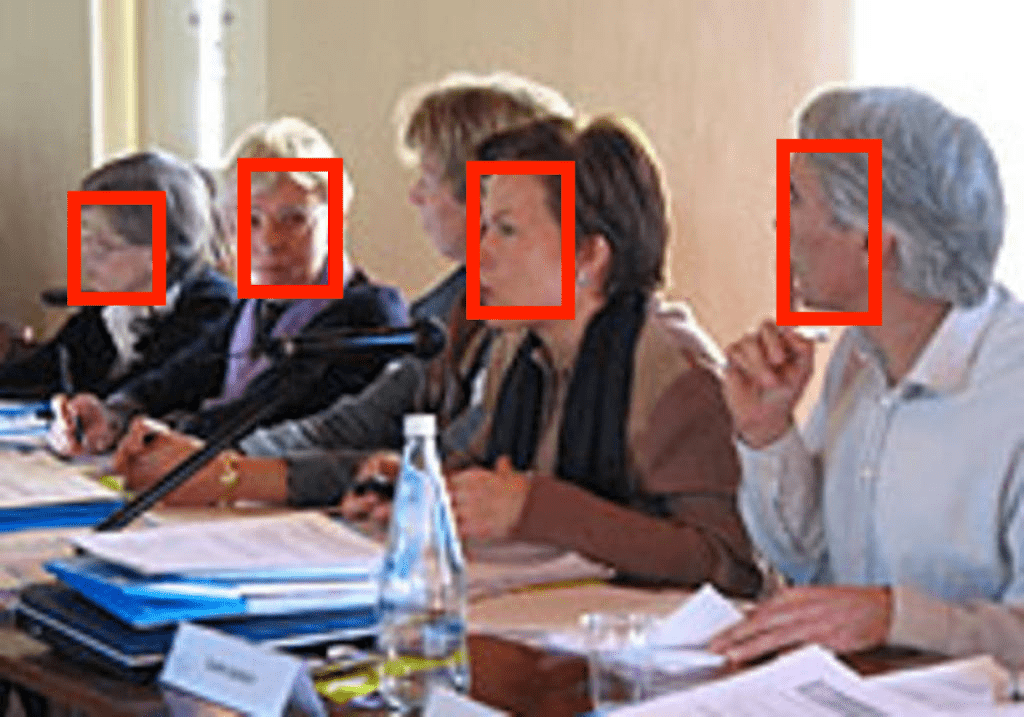} 
    ~
    \includegraphics[width = 0.2\linewidth, height = 2cm]{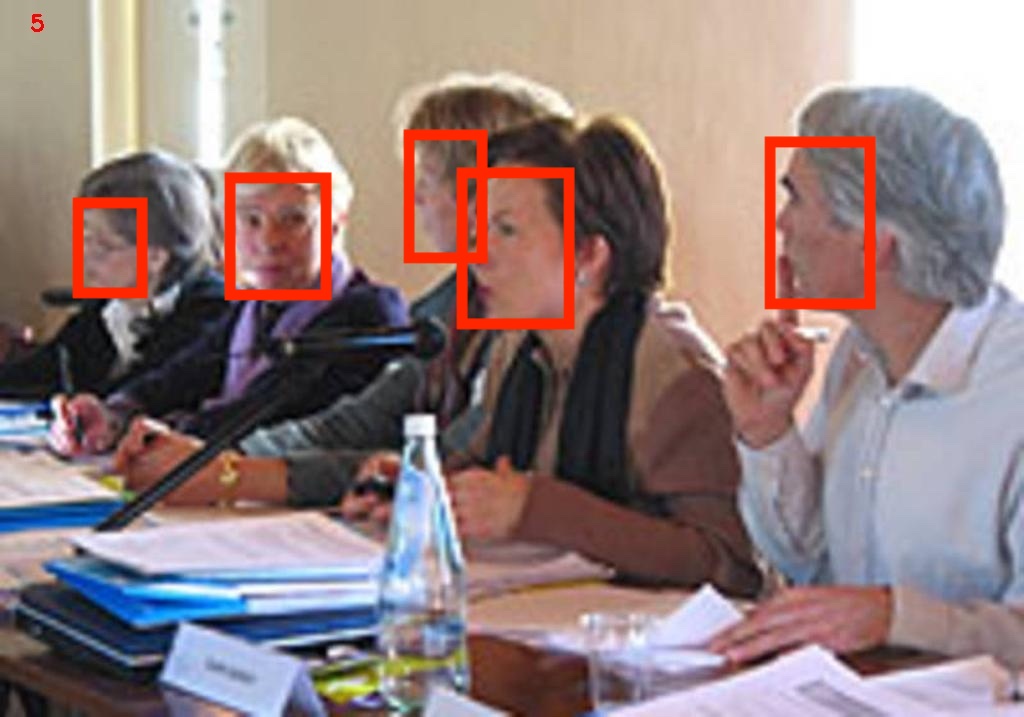}}  
    \subfigure[Reflected faces without masks]{\label{fig:fig5_reflected}
    \includegraphics[width = 0.2\linewidth, height = 2cm]{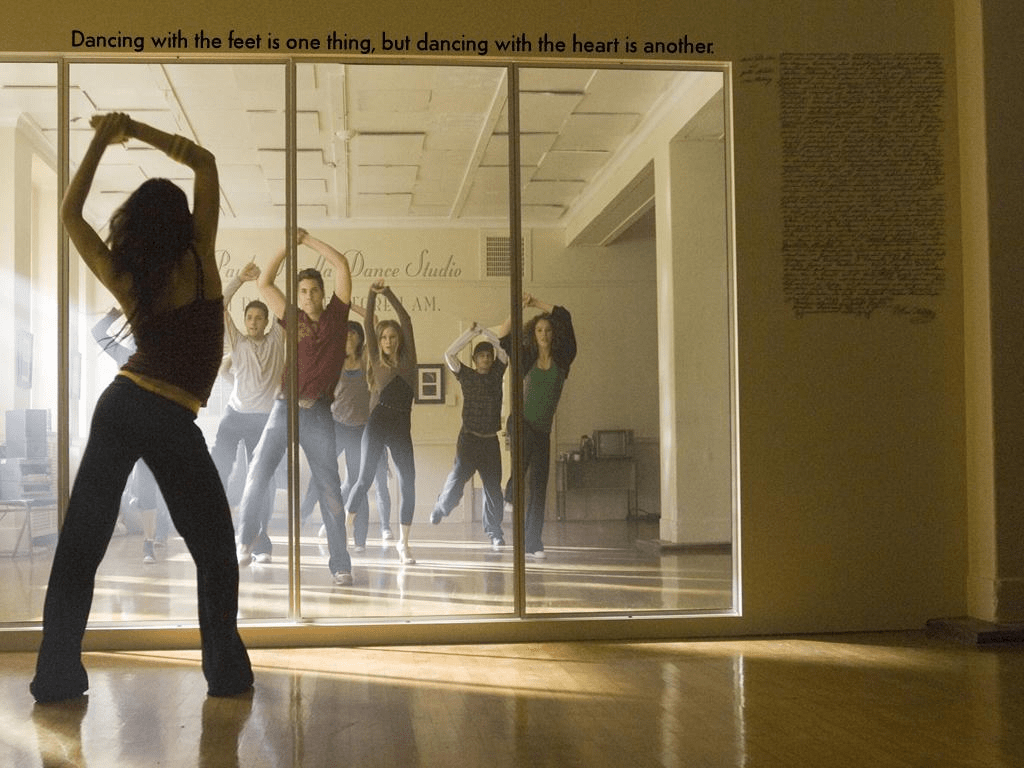}
    ~
    \includegraphics[width = 0.2\linewidth, height = 2cm]{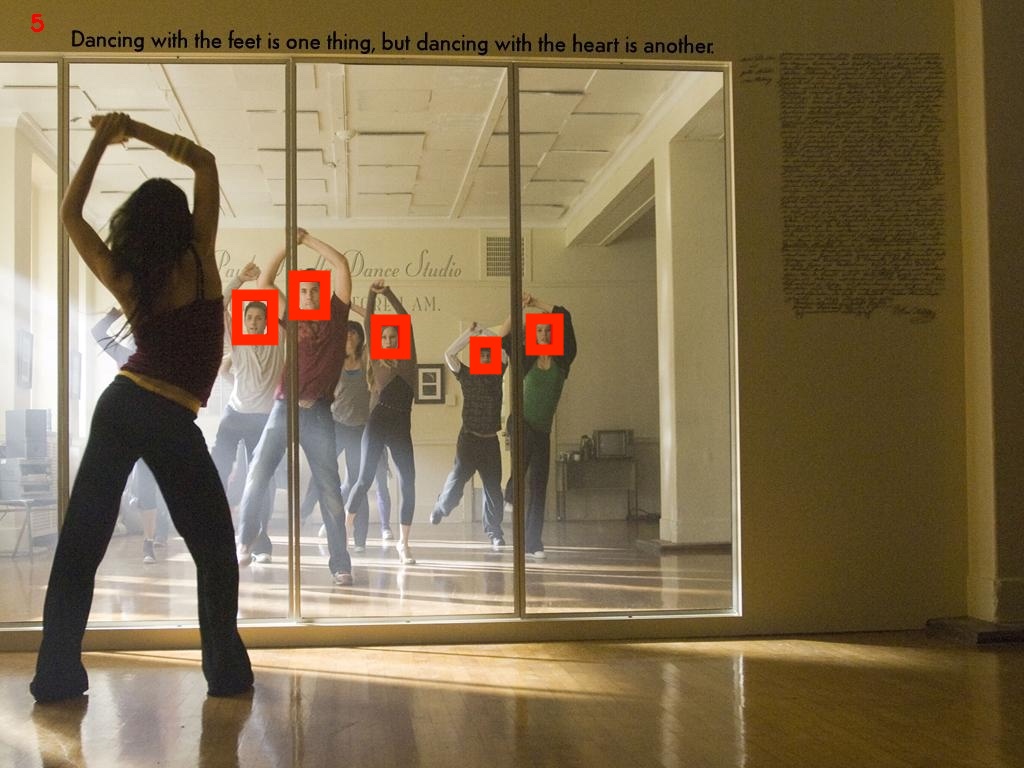}}  
    \subfigure[Faces of different orientations]{\label{fig:fig5_flipped}
    \includegraphics[width = 0.2\linewidth, height = 2cm]{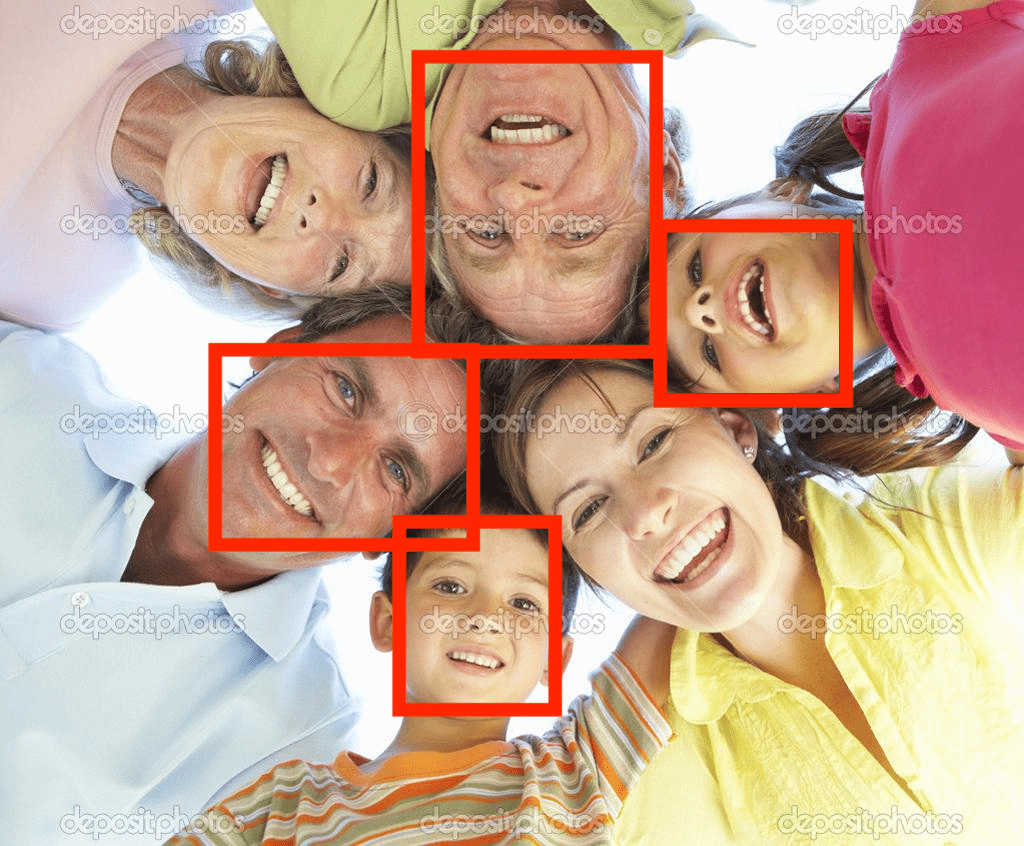}
    ~
    \includegraphics[width=0.2\linewidth, height=2cm]{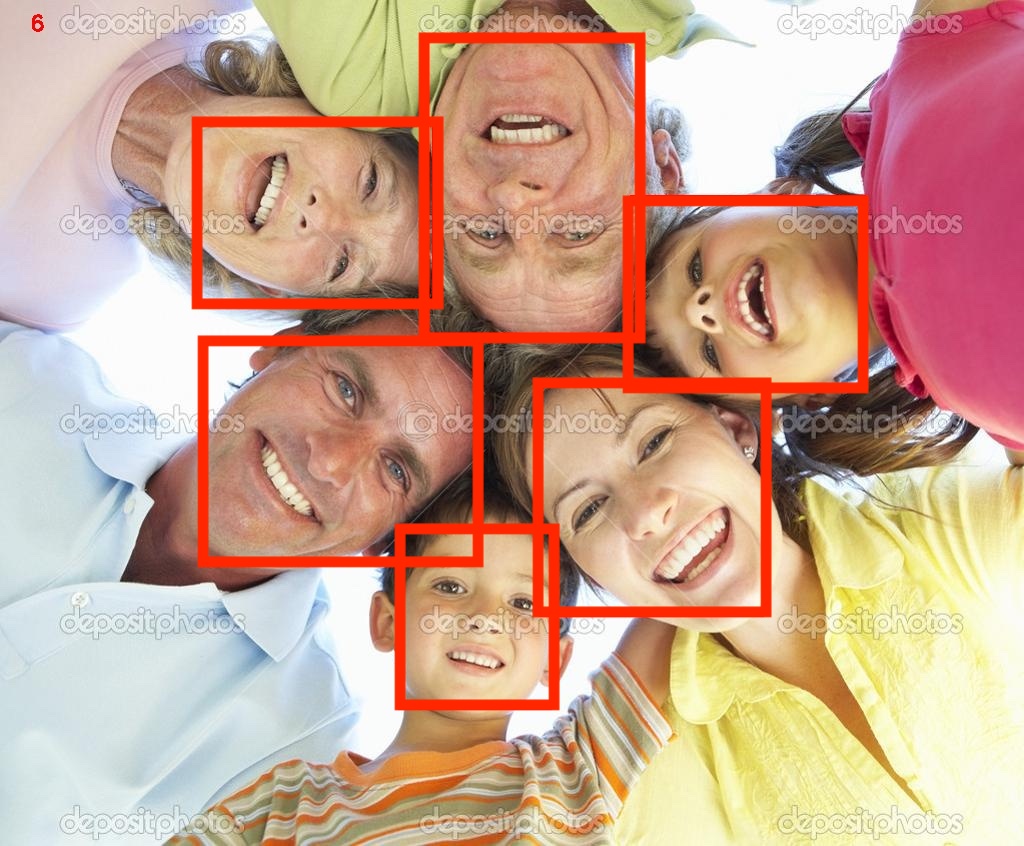}}
    \subfigure[Small faces in the meeting scene]{\label{fig:fig5_meetings}
    \includegraphics[width = 0.2\linewidth, height = 2cm]{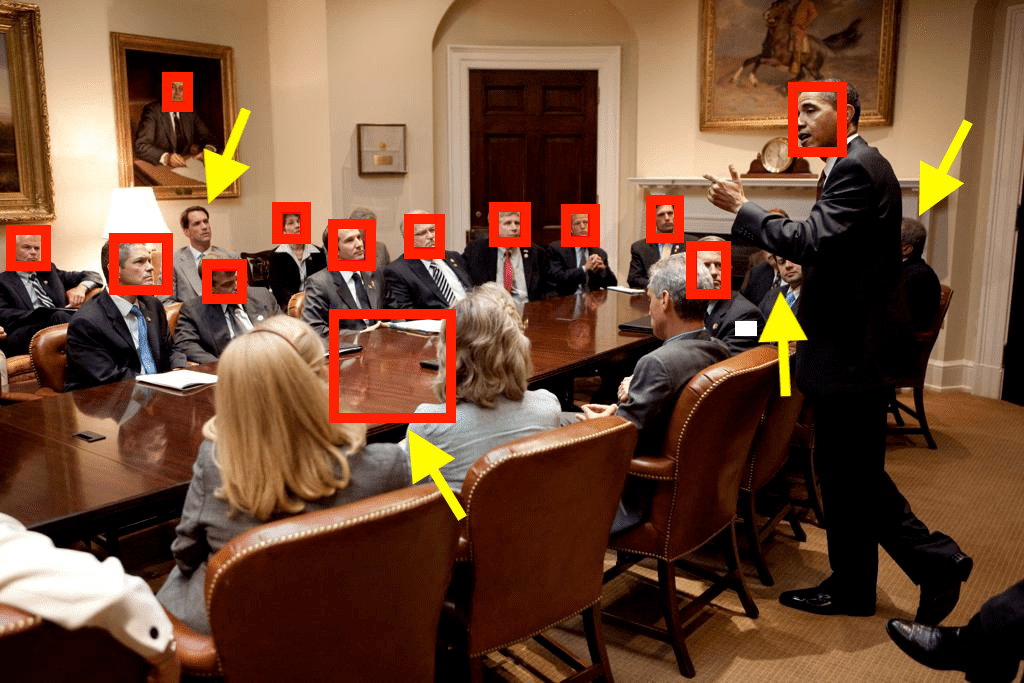}
    ~
    \includegraphics[width=0.2\linewidth, height=2cm]{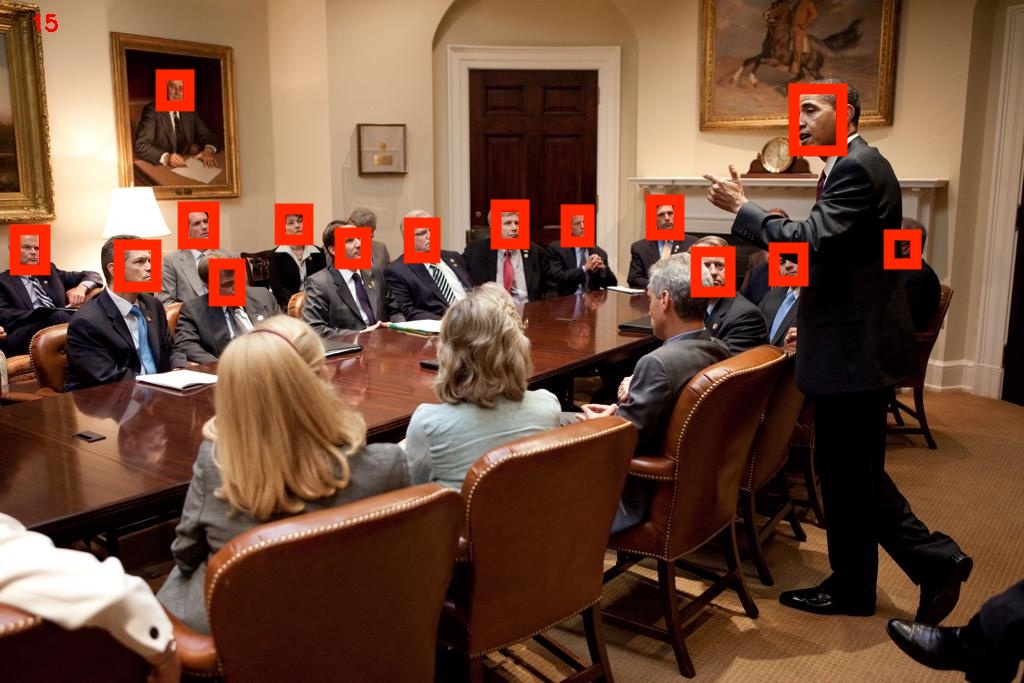}}
    \subfigure[Unlit faces]{\label{fig:fig5_unlit}
    \includegraphics[width = 0.2\linewidth, height = 2cm]{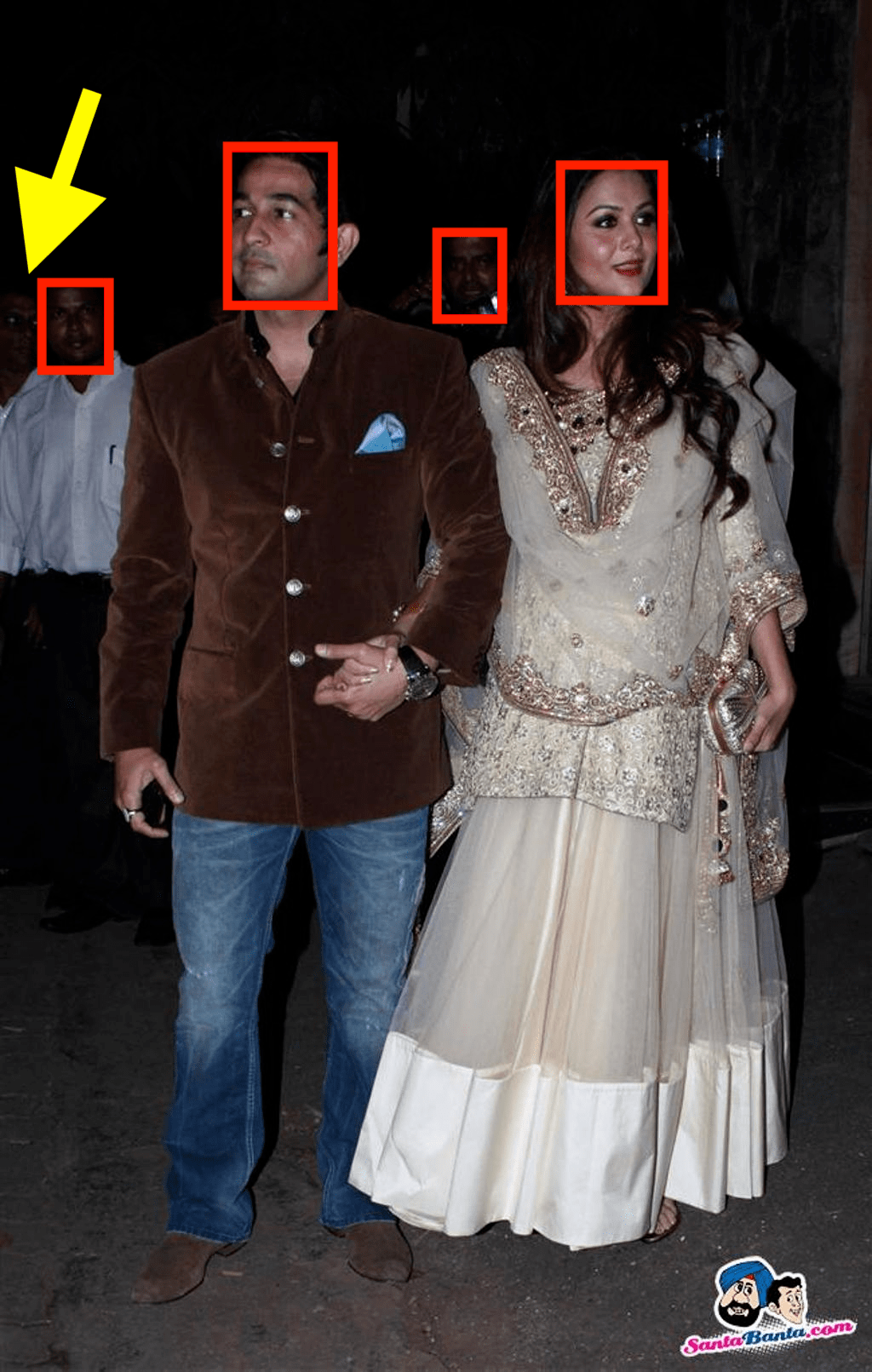} 
    ~
    \includegraphics[width=0.2\linewidth, height=2cm]{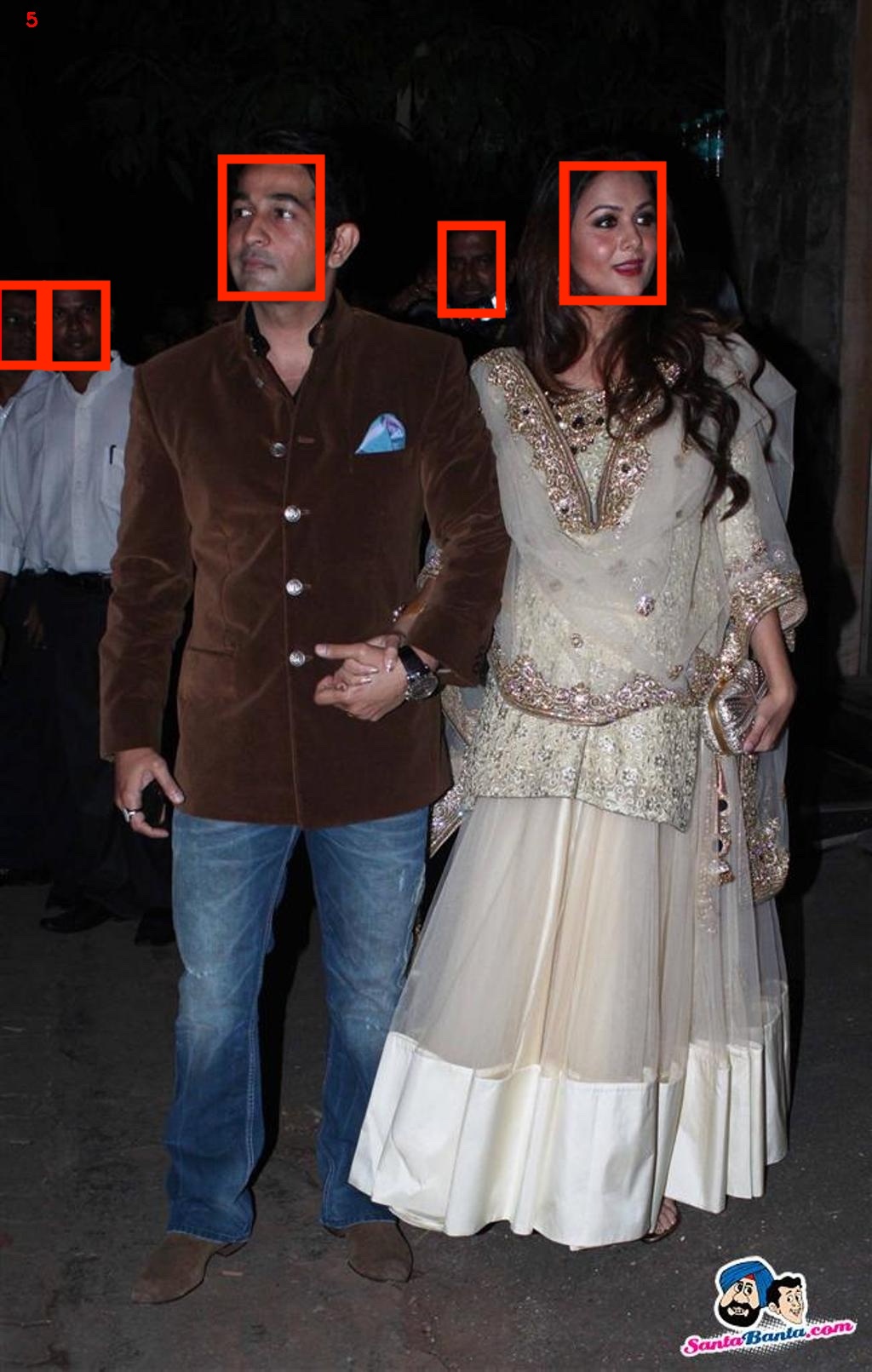}}
    \subfigure[Makeup faces without masks]{\label{fig:fig5_makeup}
    \includegraphics[width = 0.2\linewidth, height = 2cm]{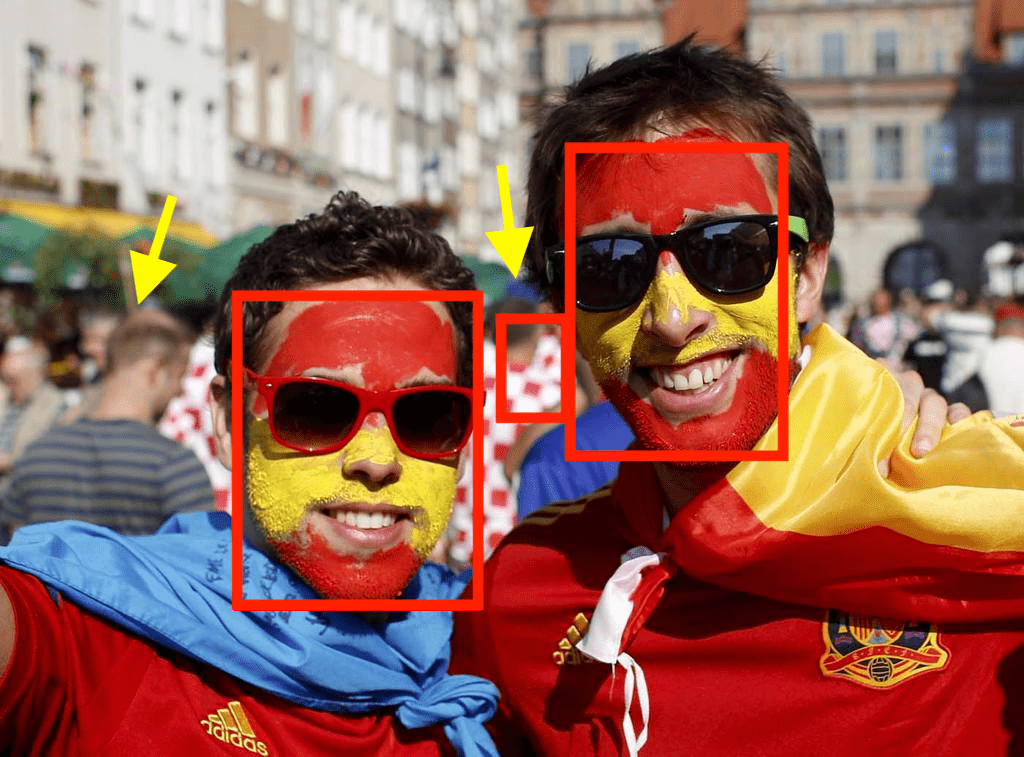} 
    ~
    \includegraphics[width=0.2\linewidth, height=2cm]{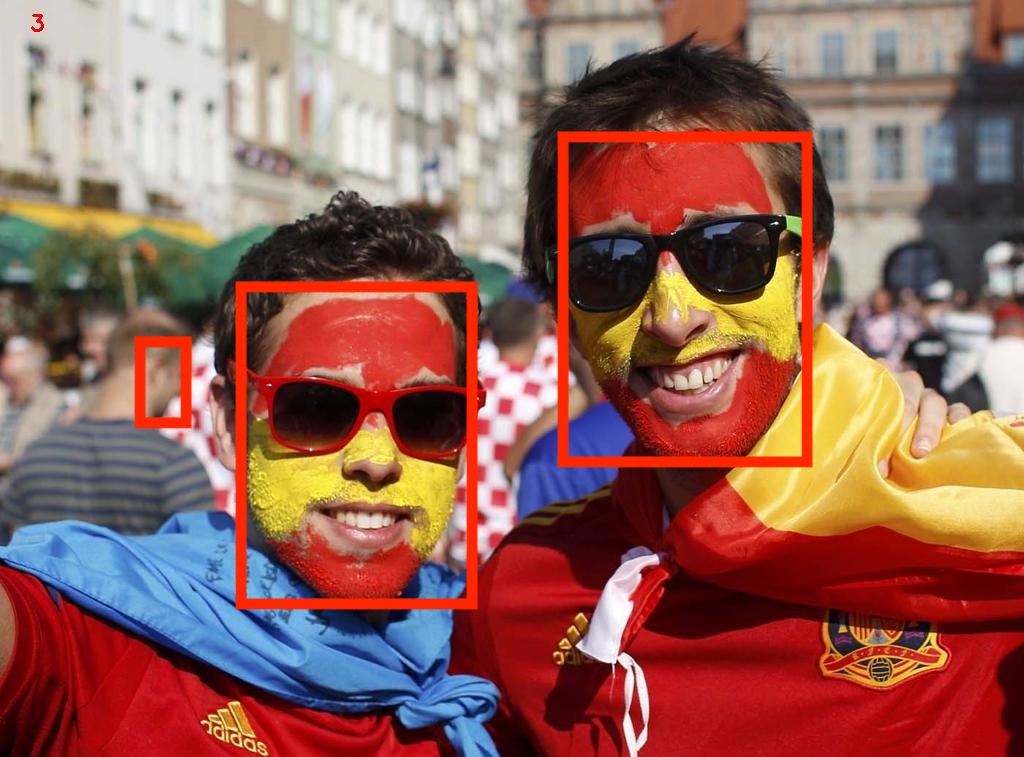}}
    \caption{Demonstrations of the detection results on the target image on the Face class, where the results of the baseline and ours are shown on the left-hand side and right-hand side, respectively. Best views in color.}
    \label{fig5} 
\end{figure}

\begin{figure}
    \centering
    \subfigure[Faces with different poses]{\label{fig:fig6_crowd}
    \includegraphics[width = 0.2\linewidth, height = 2cm]{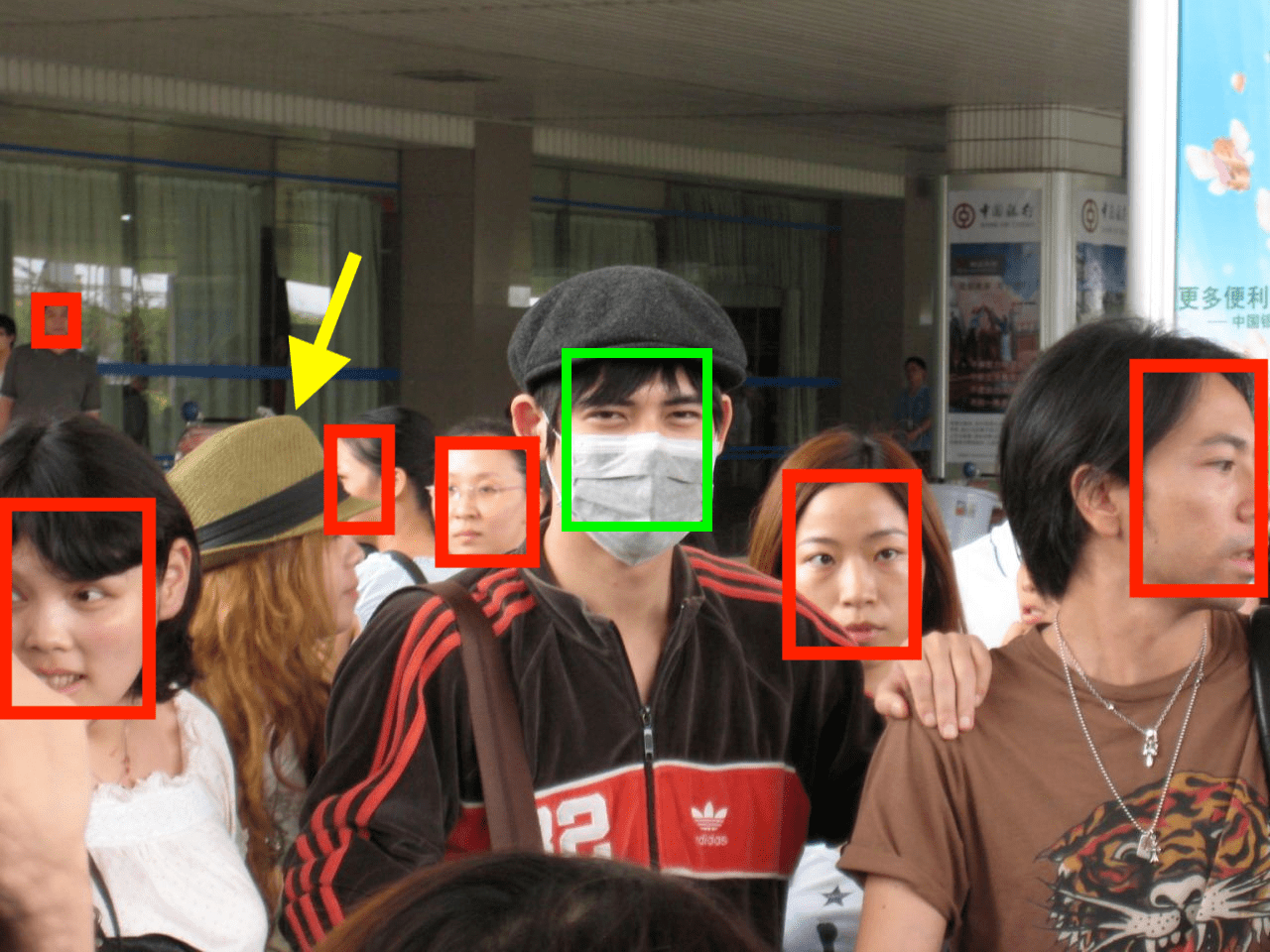} 
    ~
    \includegraphics[width = 0.2\linewidth, height = 2cm]{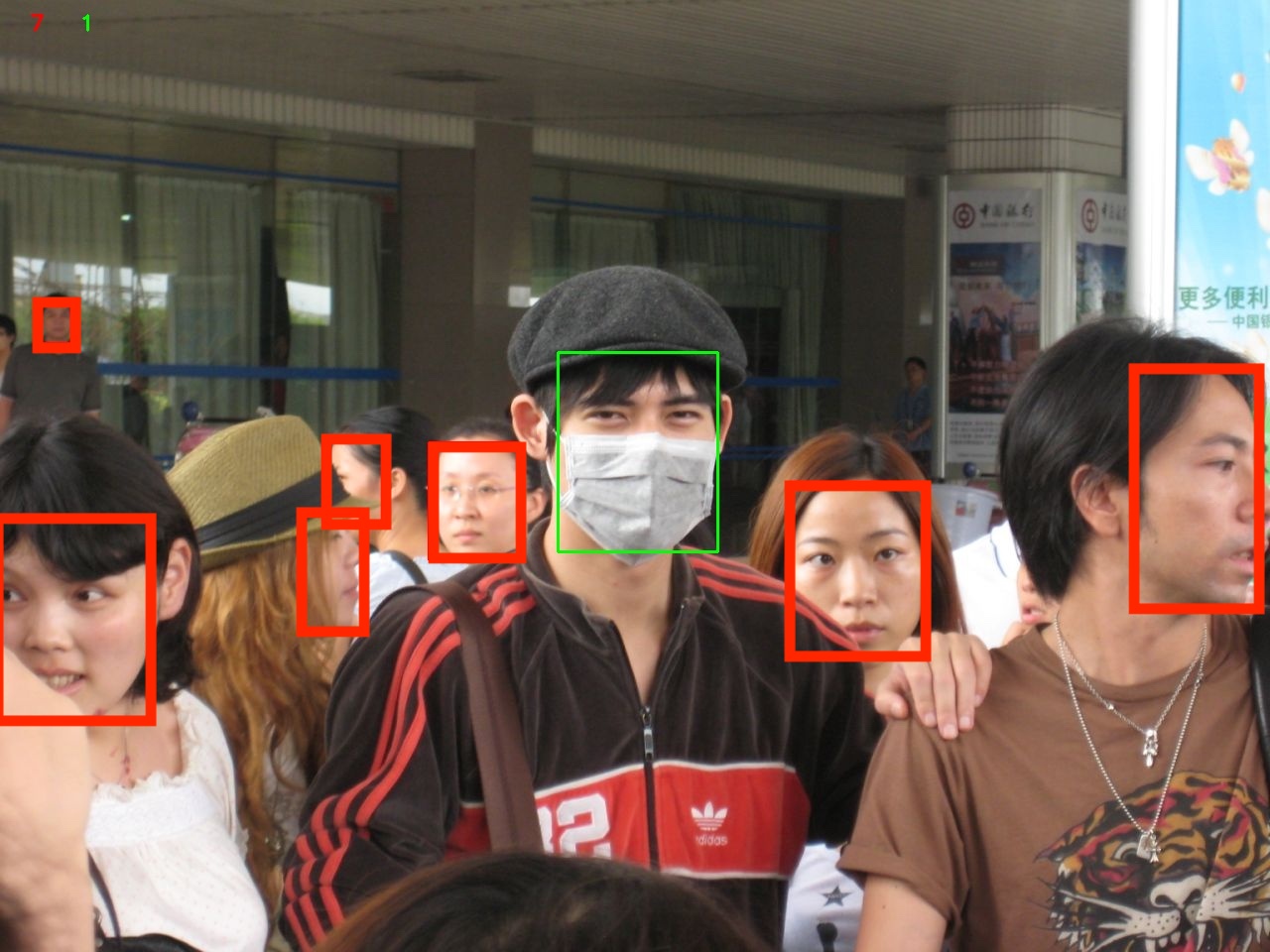}}
    \subfigure[Masks with occlusion]{\label{fig:fig6_flipped}
    \includegraphics[width = 0.2\linewidth, height = 2cm]{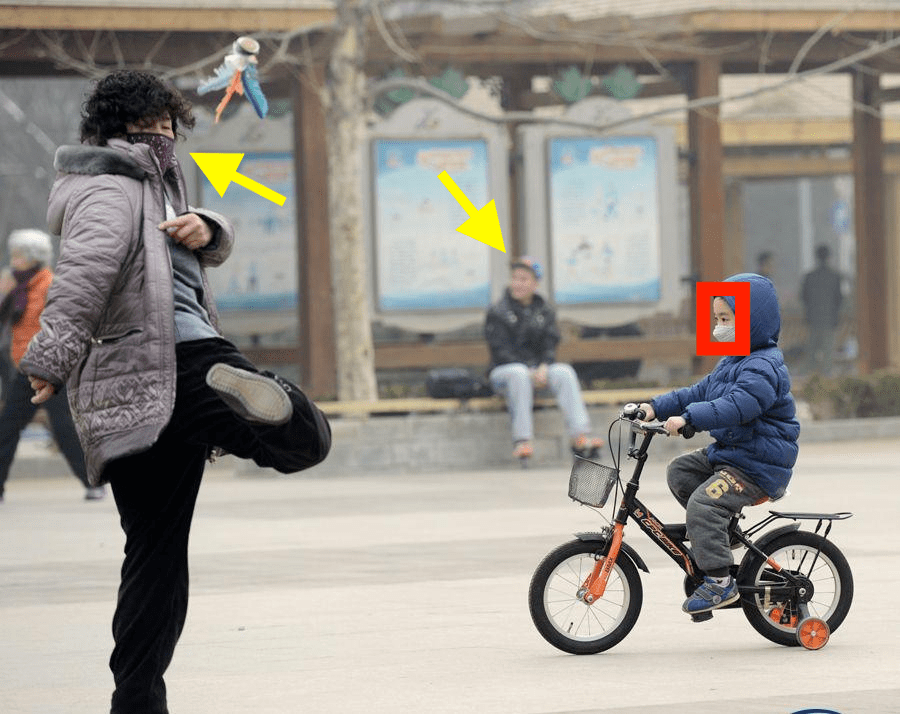} 
    ~
    \includegraphics[width = 0.2\linewidth, height = 2cm]{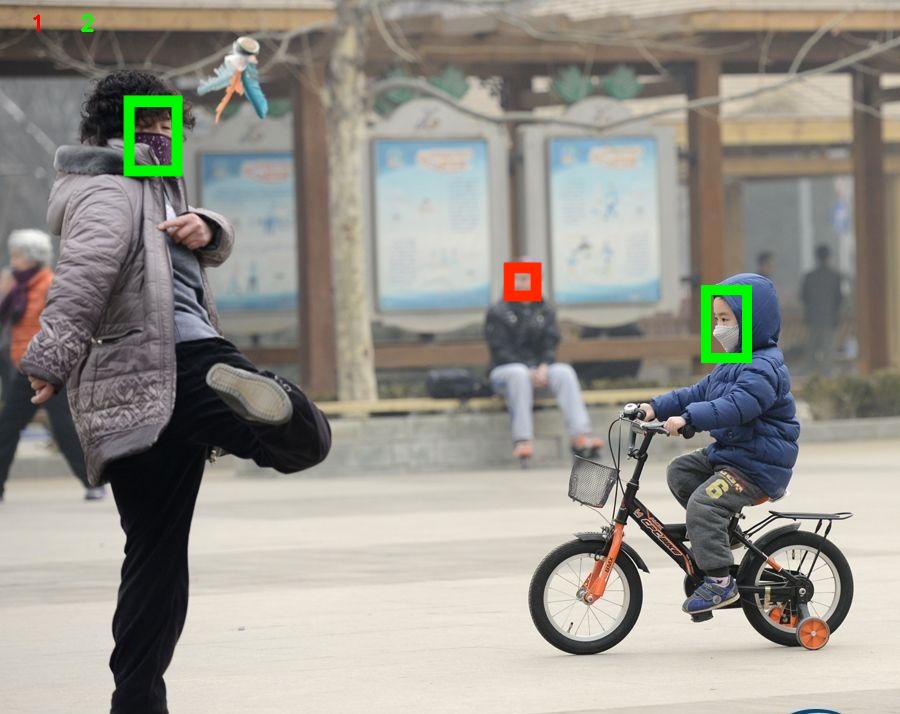}}
    \subfigure[The street view scene with and without masks]{\label{fig:fig6_angle}
    \includegraphics[width = 0.2\linewidth, height = 2cm]{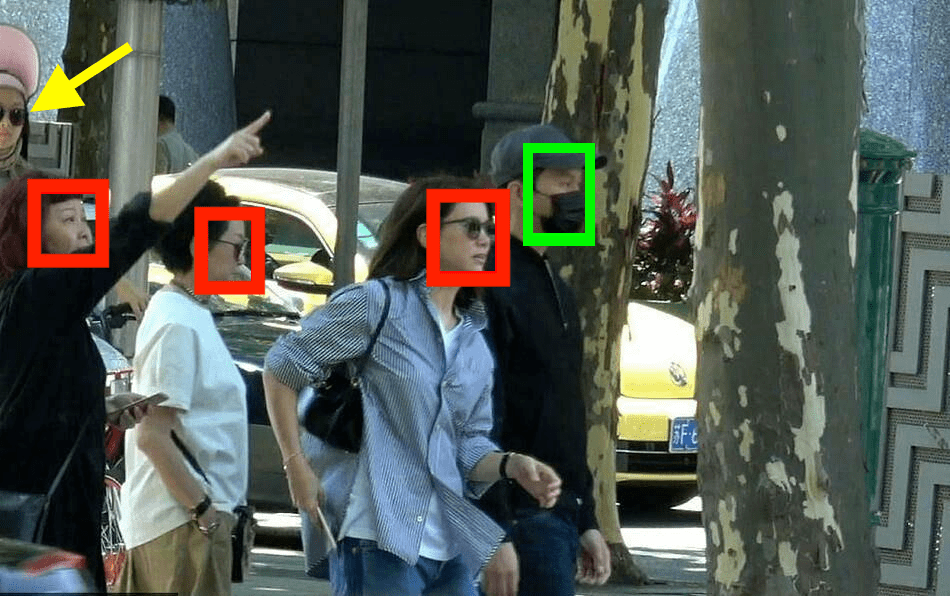} 
    ~
    \includegraphics[width = 0.2\linewidth, height = 2cm]{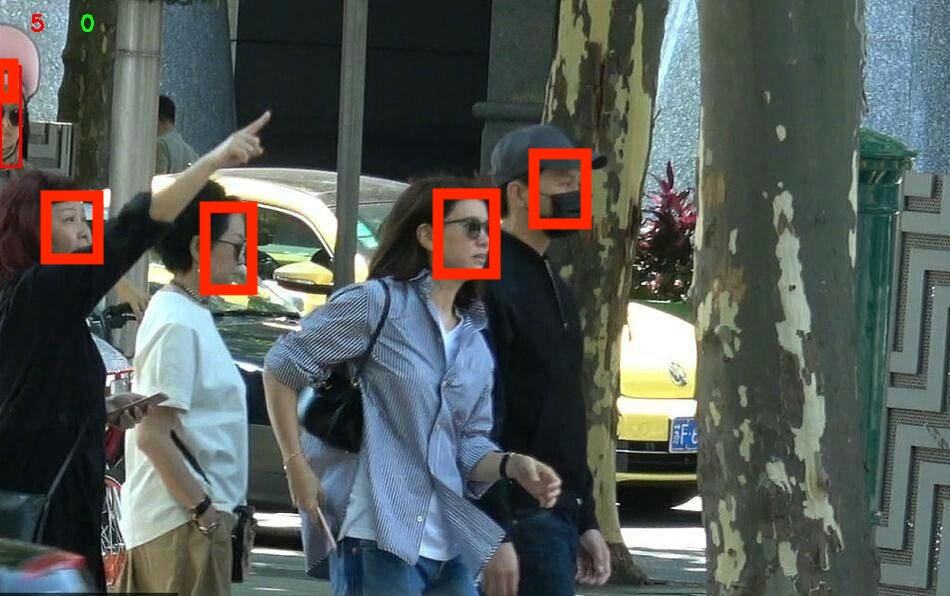}}
    \subfigure[Occlusion and masks of different colors]{\label{fig:fig6_confusing}
    \includegraphics[width = 0.2\linewidth, height = 2cm]{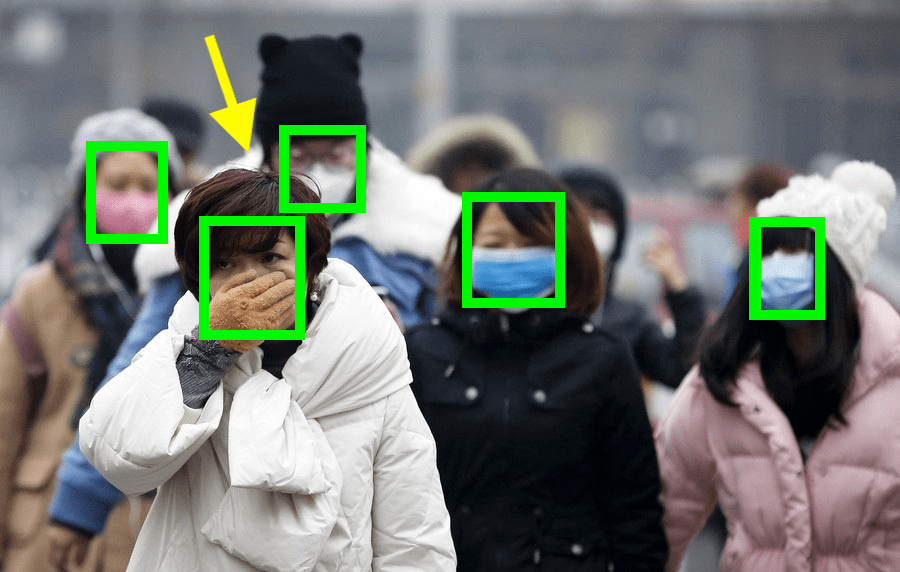}
    ~
    \includegraphics[width = 0.2\linewidth, height = 2cm]{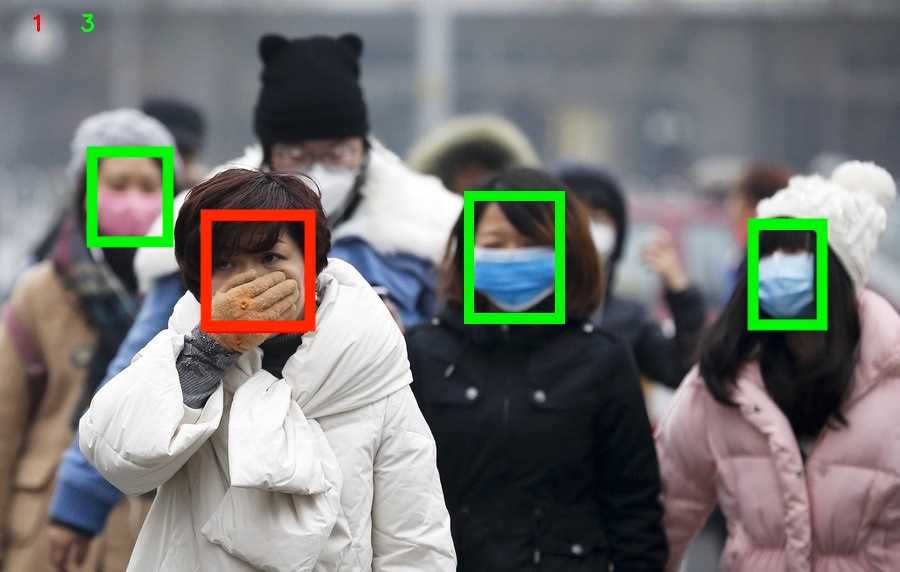}}
    \subfigure[Selfie faces]{\label{fig:fig6_blurry}
    \includegraphics[width = 0.2\linewidth, height = 2cm]{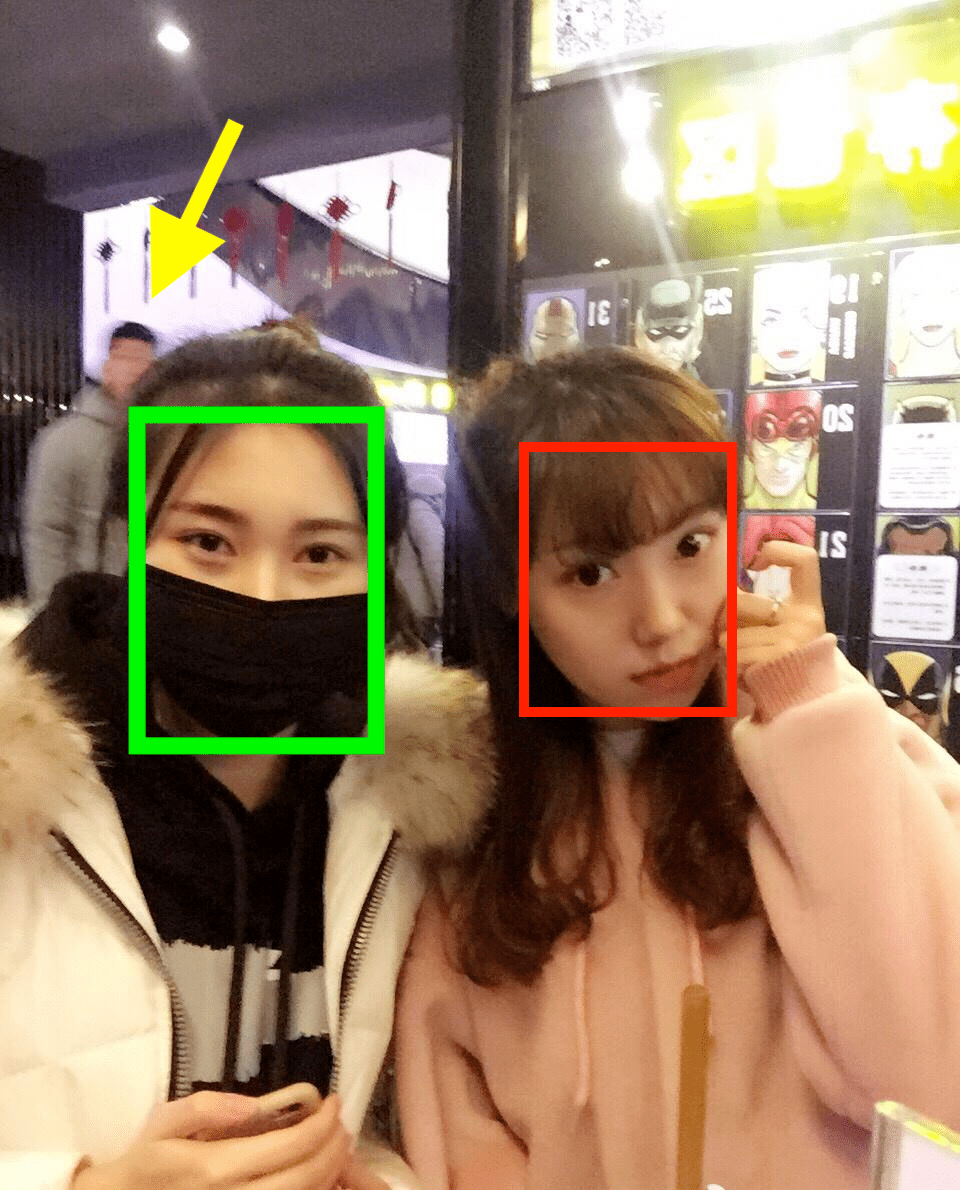} 
    ~
    \includegraphics[width = 0.2\linewidth, height = 2cm]{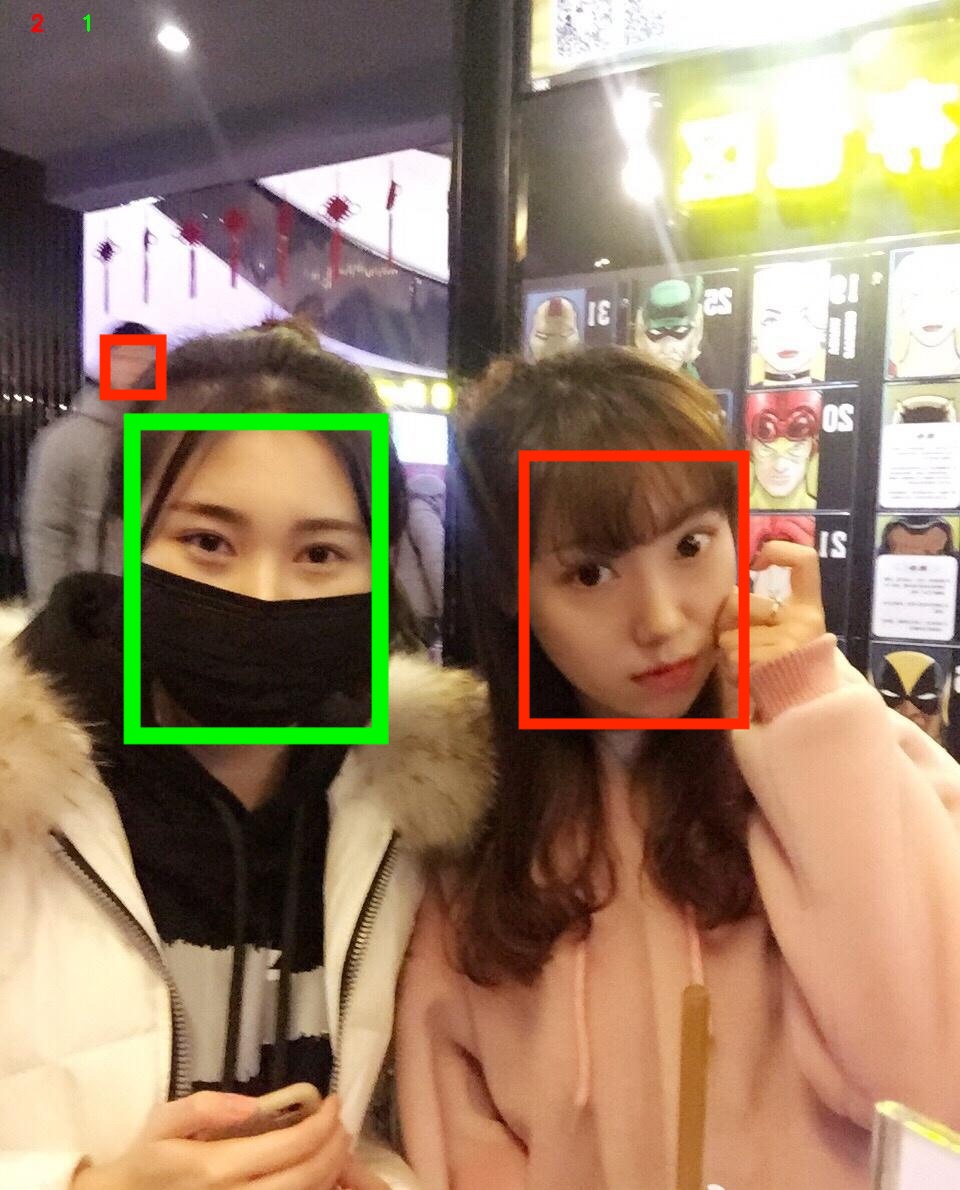}}  
    \subfigure[An obscure faces by thing]{\label{fig:fig6_hidden}
    \includegraphics[width = 0.2\linewidth, height = 2cm]{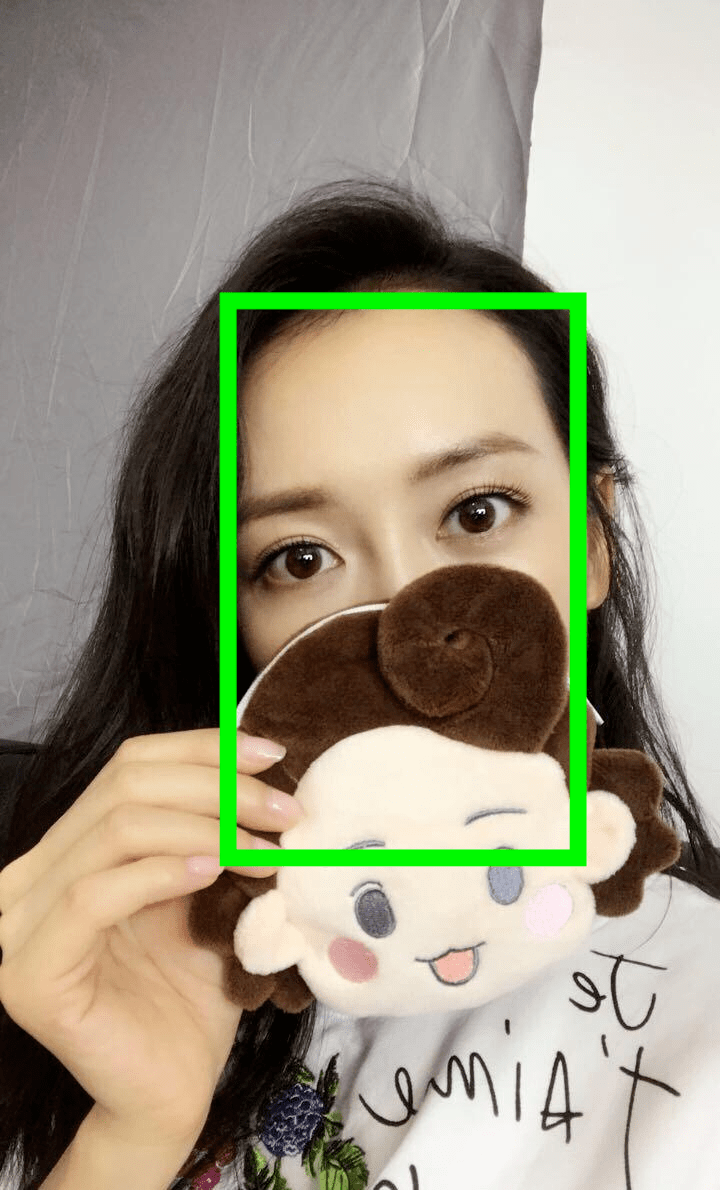}
    ~
    \includegraphics[width = 0.2\linewidth, height = 2cm]{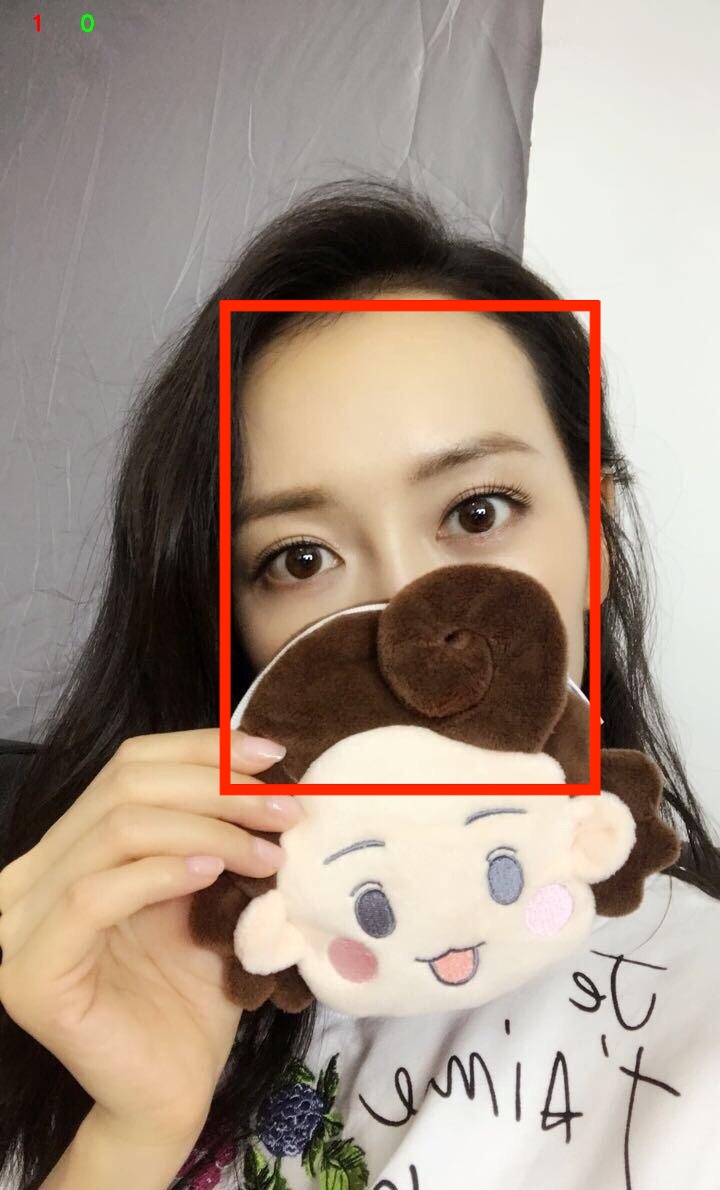}}
    \subfigure[An obscure face by a human]{\label{fig:fig6_unlit}
    \includegraphics[width = 0.2\linewidth, height = 2cm]{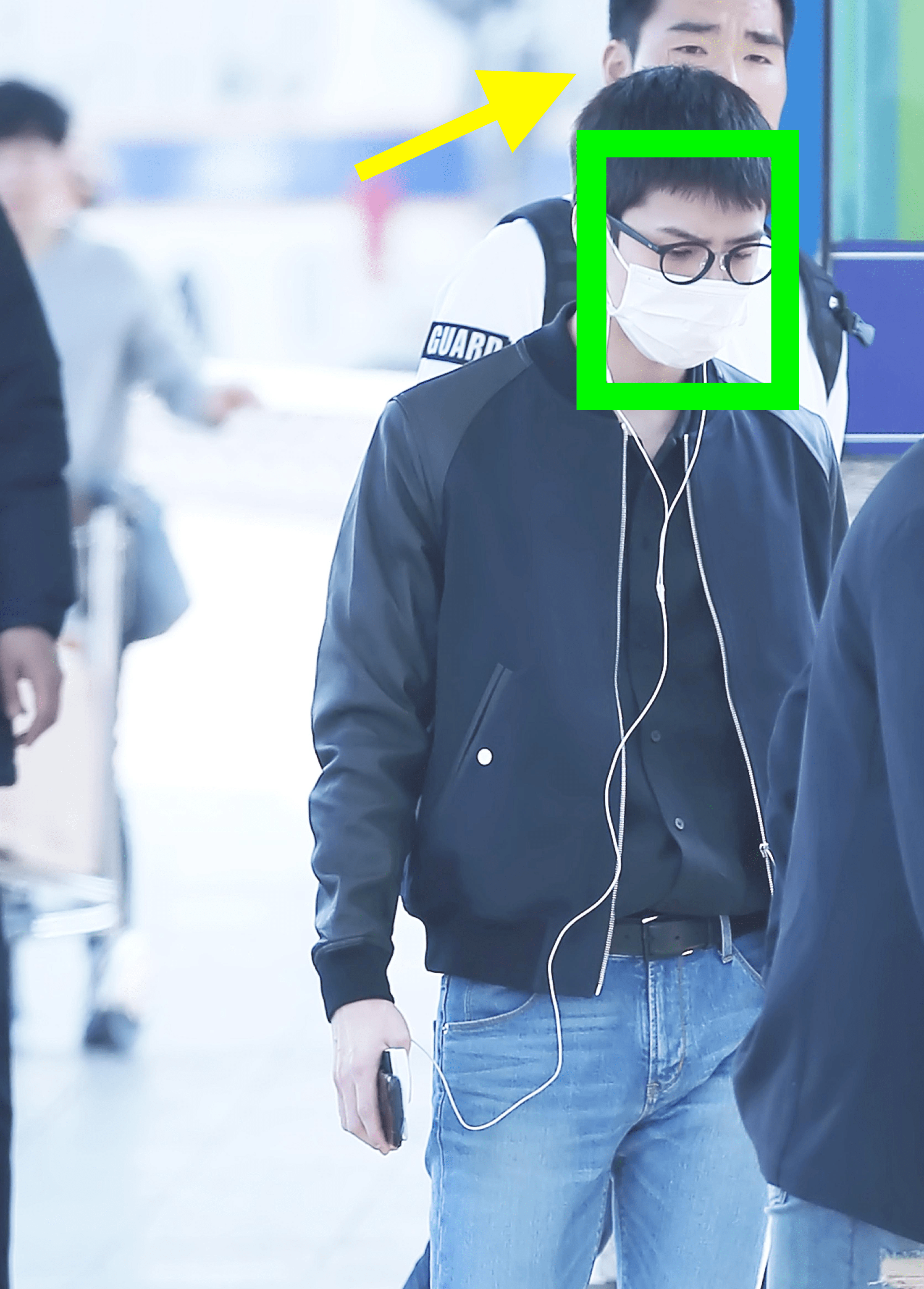}
    ~
    \includegraphics[width = 0.2\linewidth, height = 2cm]{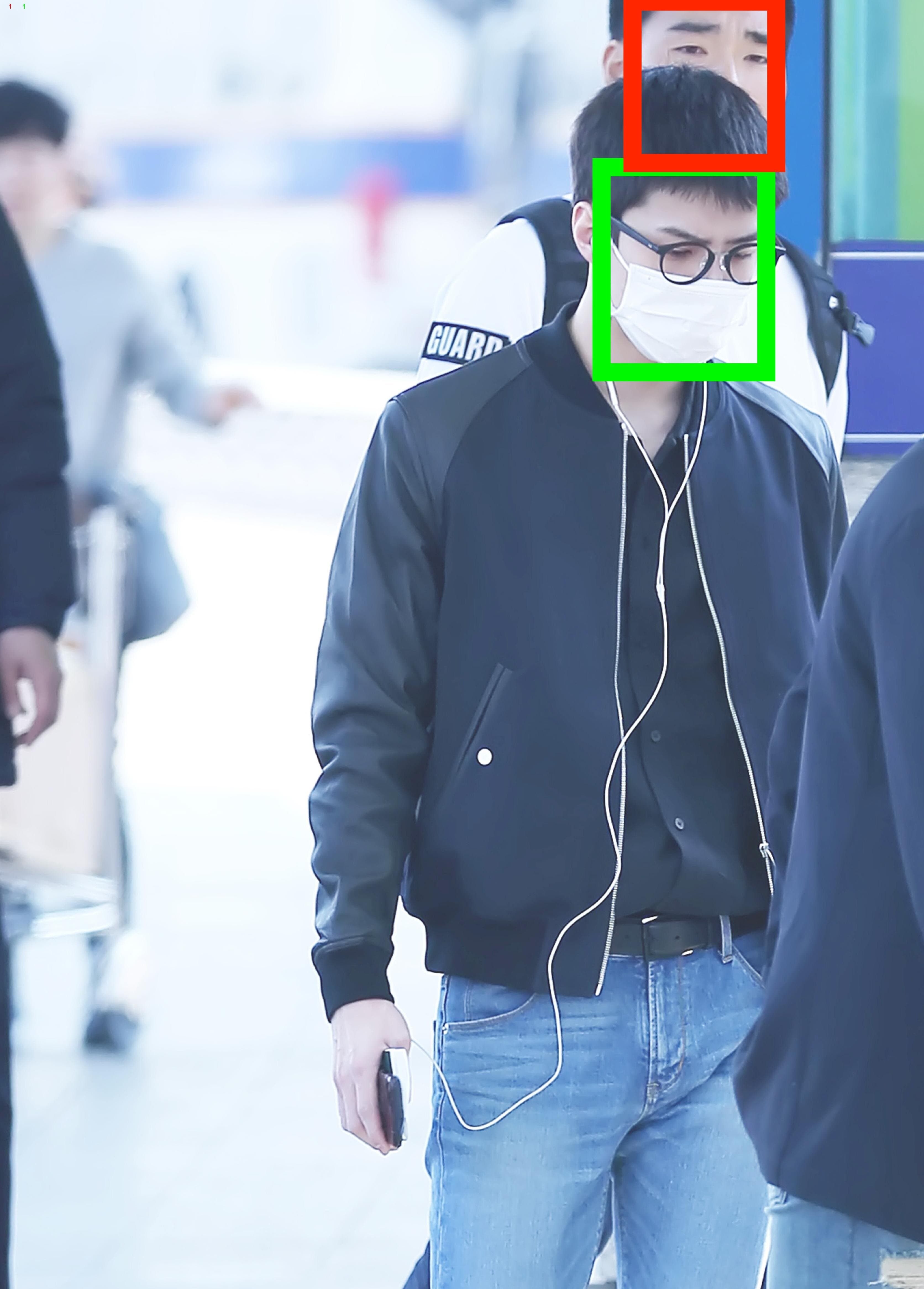}}
    \subfigure[An unclear face]{\label{fig:fig6_obsecure}
    \includegraphics[width = 0.2\linewidth, height = 2cm]{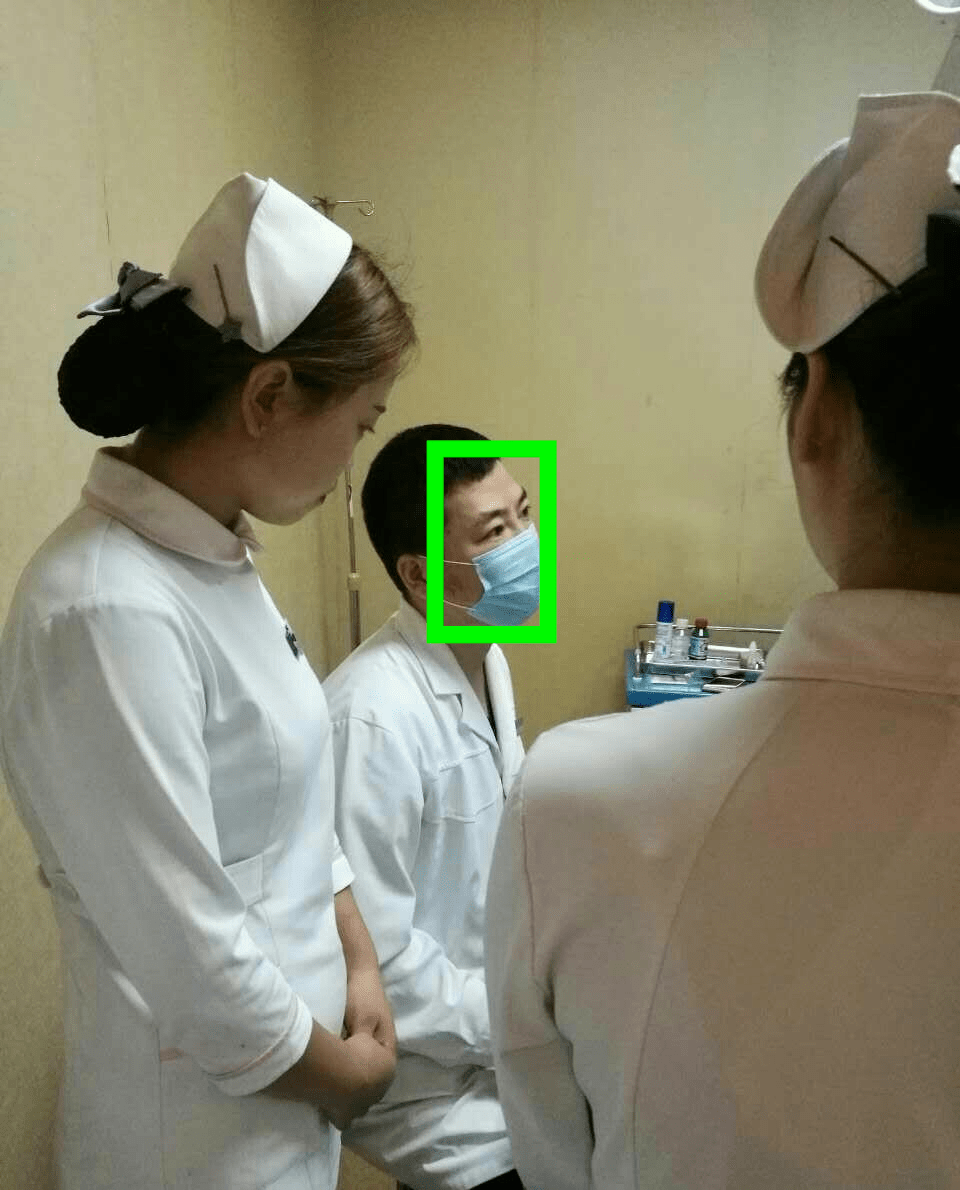}
    ~
    \includegraphics[width = 0.2\linewidth, height = 2cm]{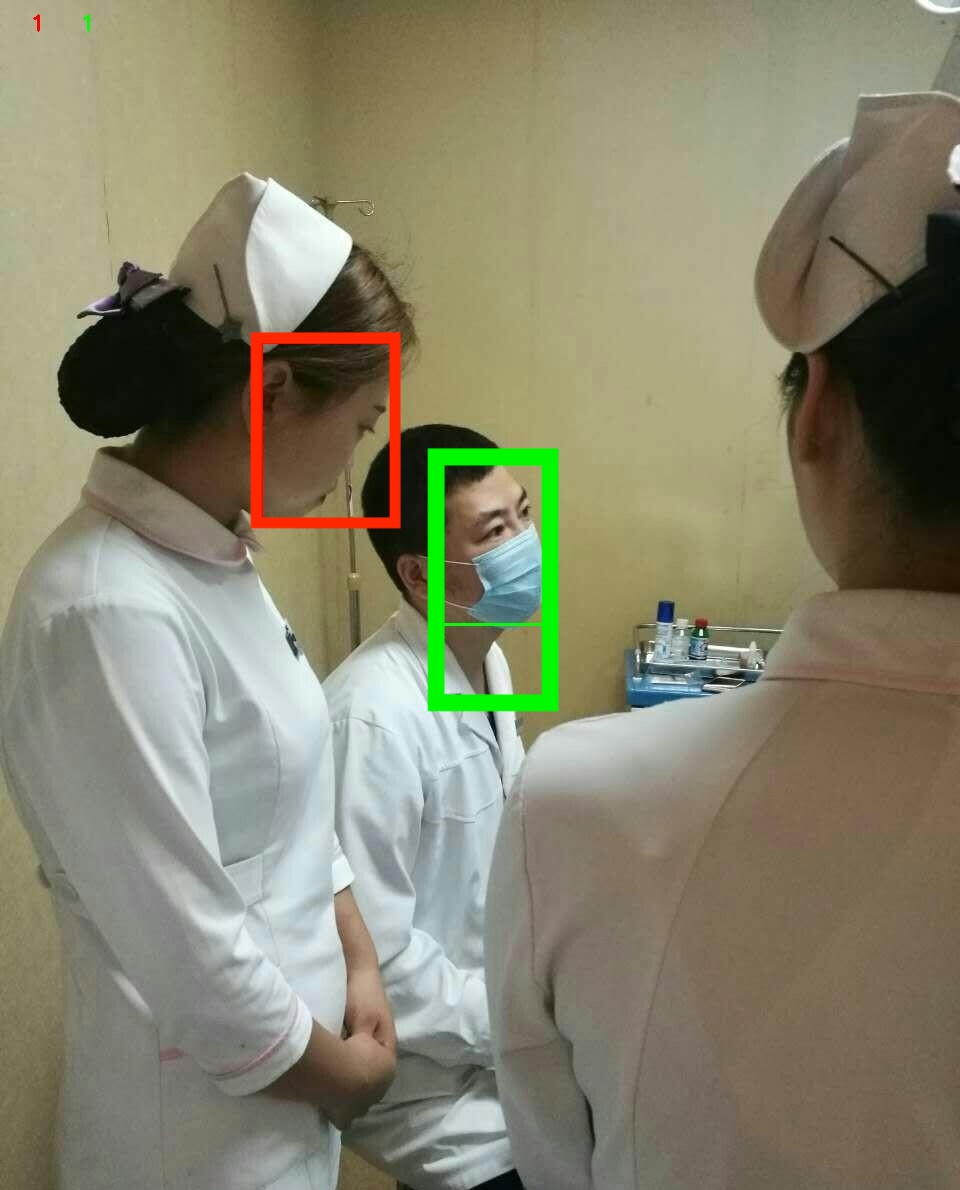}}
    \caption{Demonstrations of the detection results on the target image on the Mask class, where the results of the baseline and ours are shown on the left-hand side and right-hand side, respectively. Best views in color.}
    \label{fig6} 
\end{figure}

\noindent\textbf{Analysis on Qualitative Results.} Fig.~\ref{fig5} demonstrates the qualitative detection results of our methods on the face class, as well as the results of the baseline. The demonstration cases show the difficulties and challenges of face detection such as intra-class variation, occlusion, and multi-scale detection \cite{zou2019object}. For example, the intra-class variation of human face may present as a variety of expressions, movements, poses, and skin colors as shown in Figs.~\ref{fig:fig5_hiddens}, \ref{fig:fig5_flipped}, and \ref{fig:fig5_unlit}. Moreover, the faces may be occluded by other things as shown in Figs.~\ref{fig:fig5_reflected} and \ref{fig:fig5_makeup} or in a large variety of scales as shown in Figs.~\ref{fig:fig5_crowd}, \ref{fig:fig5_poses}, and \ref{fig:fig5_meetings}. Although these references are from various perspectives, our model is still able to extract the salient parts, while the baselines may miss one or two faces. Fig.~\ref{fig6} further demonstrates the qualitative detection results of our methods on the mask class, as well as the results of the baseline. In details, the qualitative results of the baseline are worse than that of the proposed method. For example, in various cases such as occluded faces in Figs.~\ref{fig:fig6_confusing}, \ref{fig:fig6_hidden} \ref{fig:fig6_unlit}, the intra-class issues such as various of expression, blur face, small face as shown in the rest of Fig.~\ref{fig6}. 

Due to the lack of public video benchmarks, we collect the online video clips with face masks and also conduct a laboratory experiment. For the qualitative video results, please refer to the following URLs in the footnotes.\footnote{Example 1: \url{https://drive.google.com/file/d/1Tl-8iRXgxMqeM8sbSll0tU9RubccT-lc/view}.}\footnote{Example 2: \url{https://drive.google.com/file/d/1ld1Nw3-cr1ODvzEBGkKL6hUH2dVjmojQ/view}.} The quantitative results show that 5/40 people in the video clips are missed in some frames. However, most of the missing ones are in a far distance. In contrast, 10/10 people in the laboratory experiment are correctly detected. The results manifest that the proposed model is suitable for monitoring the entrance of public transports.

\begin{figure}
    \centering
    \subfigure[Flipped faces]{\label{fig:fig7_flip}
    \includegraphics[width = 0.2\linewidth, height = 2cm]{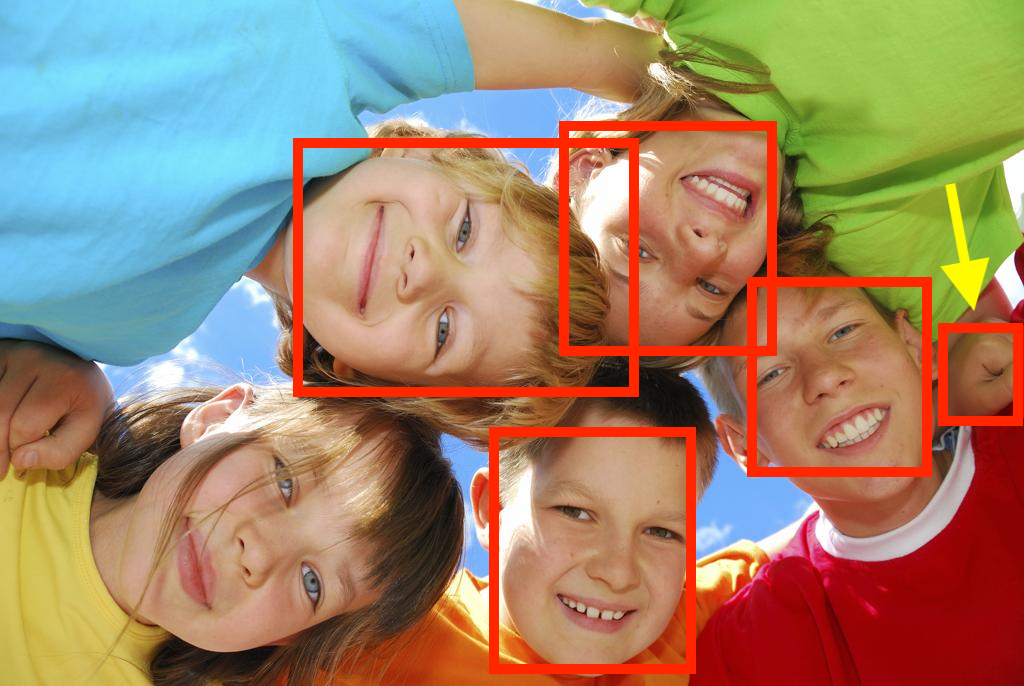}
    ~
    \includegraphics[width = 0.2\linewidth, height = 2cm]{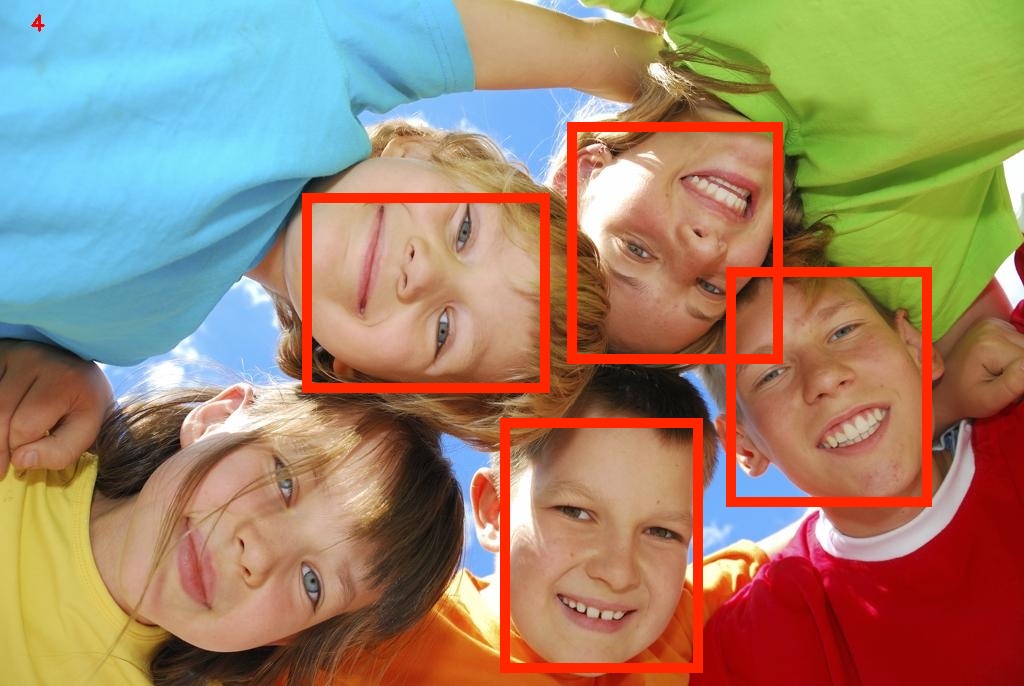}}
    \subfigure[Small faces]{\label{fig:fig7_small}
    \includegraphics[width = 0.2\linewidth, height = 2cm]{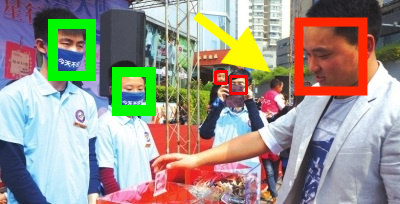}
    ~
    \includegraphics[width = 0.2\linewidth, height = 2cm]{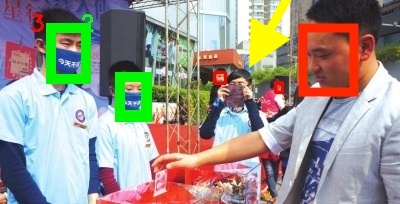}}
    \subfigure[Hidden faces by things]{\label{fig:fig7_things}
    \includegraphics[width = 0.2\linewidth, height = 2cm]{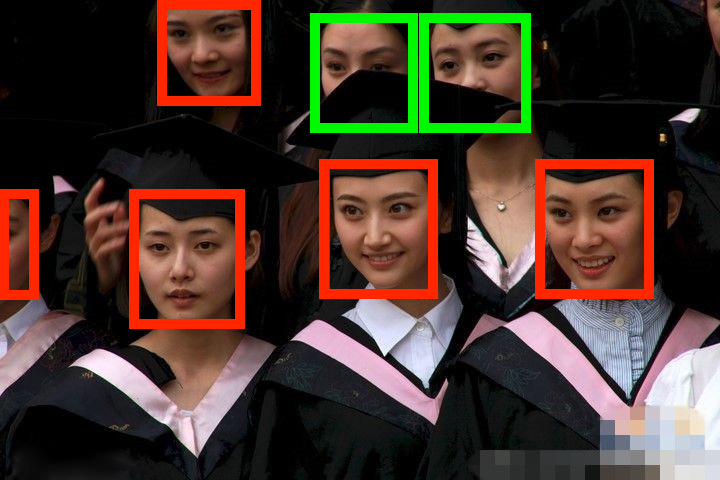}
    ~
    \includegraphics[width = 0.2\linewidth, height = 2cm]{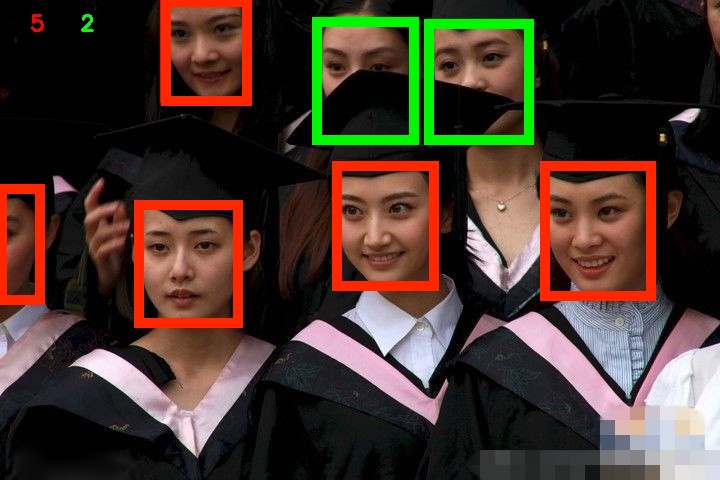}}
    \subfigure[Occlusion faces by humans]{\label{fig:fig7_hidden}
    \includegraphics[width = 0.2\linewidth, height = 2cm]{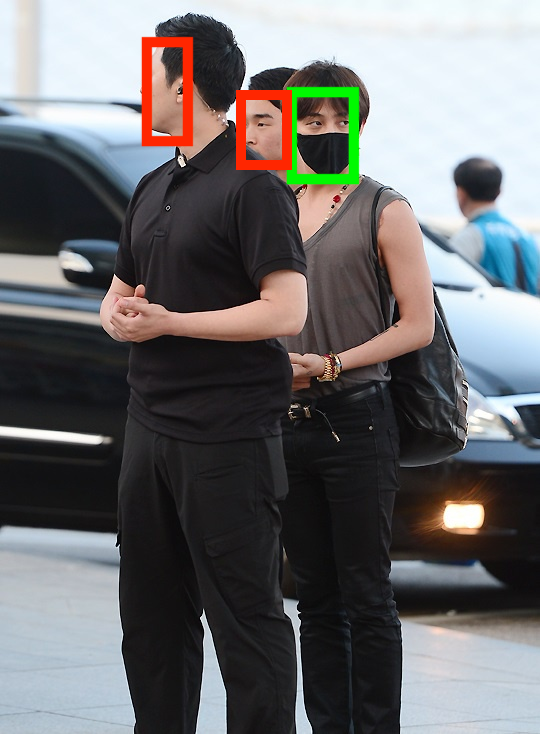}
    ~
    \includegraphics[width = 0.2\linewidth, height = 2cm]{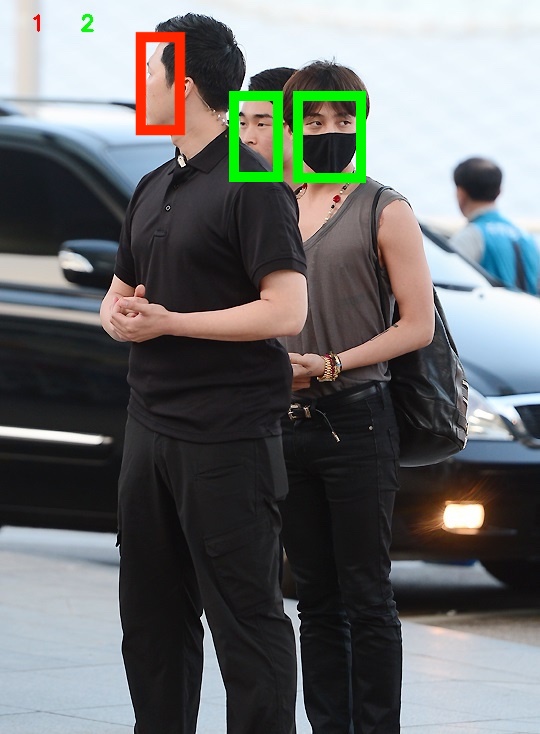}}
    \subfigure[Noisy faces]{\label{fig:fig7_noise}
    \includegraphics[width = 0.2\linewidth, height = 2cm]{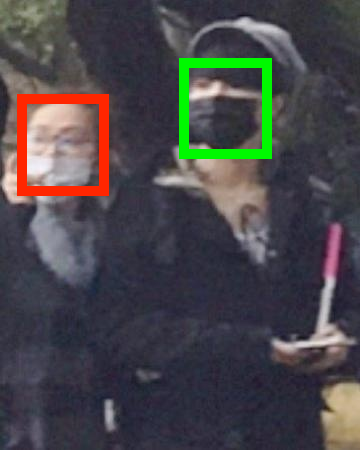}
    ~
    \includegraphics[width = 0.2\linewidth, height = 2cm]{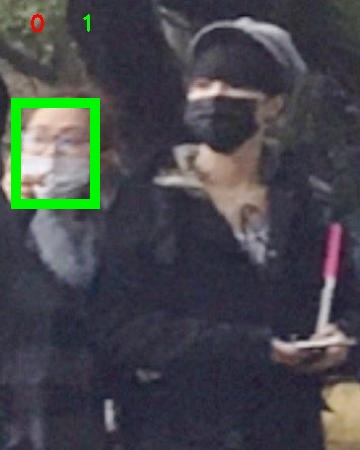}}
    \subfigure[Wrong detection faces]{\label{fig:fig7_wrong}
    \includegraphics[width = 0.2\linewidth, height = 2cm]{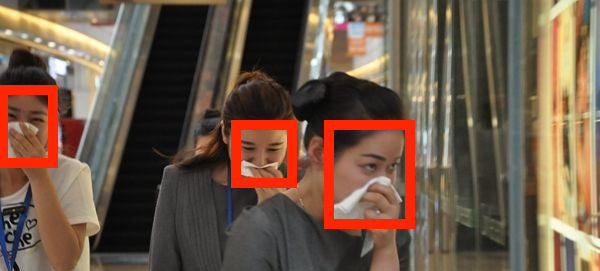}
    ~
    \includegraphics[width = 0.2\linewidth, height = 2cm]{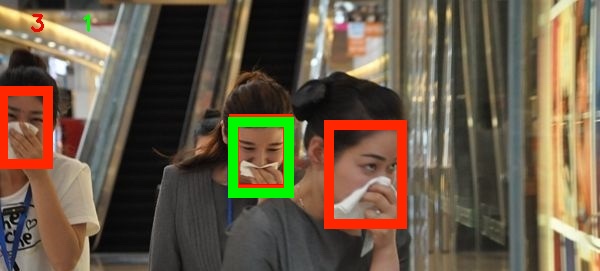}}
    \caption{Failure cases of both classes, where the results of the baseline and ours are shown on the left-hand side and right-hand side, respectively.}
    \label{fig7} 
\end{figure}

\noindent\textbf{Failure Case Analysis.} The objective loss function of "CenterFace" module is comprised of three different losses, which have different effects on the performance of the detection. For example, a face may be missed if 1) the anchor is missed, 2) imprecise embeddings relate to multiple false bounding boxes, or 3) the offset threshold is not correct. To understand how each part contributes to errors, we take a look at Fig.~\ref{fig7}. The major causes of qualitative error results happen when faces are hidden by humans or things as shown in Figs.~\ref{fig:fig7_things}, \ref{fig:fig7_hidden}, and \ref{fig:fig7_wrong}. In these cases, the detectors cannot recognize a face with or without facial masks. On the other hand, in Figs.~\ref{fig:fig7_flip}, \ref{fig:fig7_small}, and \ref{fig:fig7_noise}, faces are misrecognized when the needed detection faces are too small or in cozy environment.

\subsection{Ablation Studies}\label{abstudy}
\begin{table}
    \caption{State-of-the-art comparison}
    \label{tab:commands}
    \centering
        \begin{tabular}{ |c|c|c|c|c| }
        \hline
        \multirow{2}{*}{\bf{Model}} &\multicolumn{3}{c|}{\bf{Wider Face}}                 &\multirow{2}{*}{\bf{MAFA}} \\ \cline{2-4}
        &Easy Set &Medium Set &Hard Set &  \\ \hline
        ResNet50 \cite{jiang2020retinamask} &59.4\% &48.9\% &26.5\% &94.5\% \\ \hline
        ResNet18 &72.6\%    &71.4\%   &46.0\%  &91.5\% \\\hline
       CBAM sequence spatial-channel &19.4\%  &19.6\% &10.4\% &22.4\% \\\hline
        ResNet18 + triplet loss &76.4\%  &74.0\% &47.1\% &91.0\% \\\hline
        ResNet18 + consistency loss &74.0\% &72.2\% &45.6\% &91.8\% \\\hline
        ResNet18 + triplet loss + consistency loss &75.8\% &73.0\% &46.3\% &91.4\%\\\hline
        {CenterFace (all modules)} &{90.2\%} &{ 85.4\%} &{54.1\% } &{ 91.5\%} \\\hline\hline
        { ResNet18 + self-supervised rotation \cite{gidaris2018unsupervised}} &{ 65.5\%} &{ 63.0\%} &{ 38.3\% }&{ 91.4\% }\\\hline
        { ResNet18 + self-supervised grayscale~\cite{ssECCV2020}} &{ 60.2\%} &{ 61.4\%} &{ 39.0\%} &{ 91.5\%} \\\hline        
        { ResNet18 + self-supervised random crop~\cite{ssECCV2020}} &{ 60.3\% }&{ 58.9\%} &{ 37.8\% }&{ 91.9\%} \\\hline
        \end{tabular}
        \label{table:state}
\end{table}
Here, the ablation studies are presented to show the effectiveness of each component in the proposed framework. We perform the ablation studies on the effectiveness of objective function and training setting. We use WiderFace and MAFA datasets to show the results. We compare with the recently state-of-the-art detector in WiderFace and MAFA test-dev in Table~\ref{table:state}. We also adopt three objective loss functions in Eq. \ref{totalloss} to evaluate their individual performance.

Please note that the ResNet backbone here is utilized alone without the attention mechanism to certify the performance of objective loss functions. The results show that our module outperforms the baseline without adding any objective losses. Afterward, we add the proposed losses one-by-one to observe their effects on the accuracy. This study shows that the proposed objective loss functions significantly boost the performance on the WiderFace dataset.

\noindent\textbf{Attention Mechanism.} In this study, We compare the two different ways of arranging the channel and spatial attention sub-module in the convolution neural network architecture. As each module has a different setting function, the order may affect the overall performance. For example, the spatial attention works locally while the channel attention is applied globally. Indeed, in Table~\ref{table:state} our experiment shows that it yields a lower accuracy if we arrange the sub-modules channel and spatial in different ways. The reason is that in traditional architecture the channel attention focus on "what" is an informative part, e.g., the coordinate part in the feature space, which is complementary to the spatial attention. The informative parts then concatenate with the computed pooling from the spatial attention module to generate the efficient feature descriptor. the vise-versa architecture of the attention module may make the descriptor miss to learn some crucial features such as corner area or flipped angle of objects.

\noindent{ \textbf{Comparisons with Self-Supervised Learning.} In addition, we conduct another experiment for demonstrating the performance of consistency loss. Specifically, we use the other widely-used self-supervised pretext task for face mask detection, i.e., rotation~\cite{gidaris2018unsupervised}. Furthermore, following the previous work~\cite{ssECCV2020}, we also apply other augmentation methods like grayscale and random crop as the self-supervision tasks. Table~\ref{table:state} shows that all other self-supervision tasks decrease the accuracy of the framework on the WiderFace dataset while keeping a similar accuracy on the MAFA dataset. This is because of the characteristics of the dataset, i.e., the WiderFace dataset includes people in different angles and various appearance diversity while most images in the MAFA dataset are the small group or single people with or without masks. As such, knowing the knowledge from the self-supervised tasks cannot further improve or even deteriorate the performance. It is also worth noting that the objects are different when applying the self-supervised learning in the original paper. Therefore, knowing the rotation angles means that the model understands the semantics of the objects. However, the subjects in the face mask detection are almost about people. As such, knowing the rotation does not mean that the model understand the semantic information. In contrast, imposing the consistency constraint on the prediction results of original images and their horizontally-flipped images simultaneously stabilizes the prediction results and enlarges the training data. Therefore, the consistency loss facilitates the learning process.}


\begin{figure}
    \centering
    \includegraphics[height = 4.8cm]{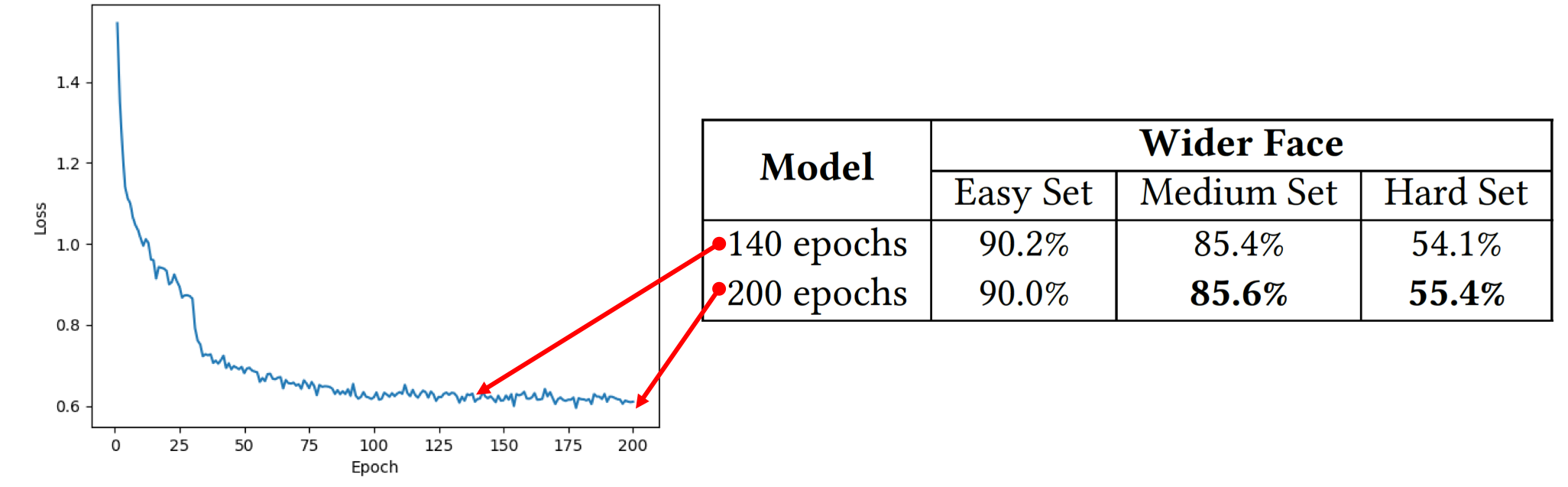} 
    \label{fig:fig8_200}
    \caption{Training curve (loss with different numbers of epochs).}
    \label{fig8} 
\end{figure}

\noindent\textbf{Training Curve.} { The default number of training epochs is 140 epochs with a learning drop at 30 and 80 epochs. We further train the model to 200 epochs to observe if there is any improvement in performance. As shown in Figure~\ref{fig8}, the results manifest that the medium and hard sets of the WiderFace dataset further increases by 0.2\% and 0.3\% while the result of MAFA dataset increases by 0.2\%. The improvement is relatively minor while the training requires much longer training time. Therefore, to keep the performance of the algorithm without wasting the computational resource, we suggest to train the the proposed model with 140 epochs.}    

\begin{table}
    \caption{Training performance comparison, where L1 loss works a bit better than Smooth L1 loss.}
    \label{tab:commands}
    \centering
        \begin{tabular}{ |c|c|c|c|c| }
        \hline
        \multirow{2}{*}{\bf{Model}} &\multicolumn{3}{c|}{\bf{Wider Face}}                 &\multirow{2}{*}{\bf{MAFA}} \\ \cline{2-4}
        &Easy Set &Medium Set &Hard Set &  \\ \hline
        Smooth L1 &90.0\%    & 85.3\%   &\bf{55.2\%}  &91.2\% \\
        L1  &{\bf{90.2\%}}  & \bf{85.4\%} & 54.1\% & \bf{91.5\%} \\ \hline
        \end{tabular}
        \label{table:loss}
\end{table}

\noindent\textbf{Regression Loss.} The regression loss is the distance loss, which enforces the distance between the anchor to the positive and negative pair. Certainly, the smooth L1 loss is usually a common choice for classification, regression, and distance loss in object detection problems. Similar with the observation from previous works~\cite{sun2018integral, sun2017compositional}, L1 loss yields better accuracy at a fine scale as compared to Smooth L1 loss \cite{law2018cornernet}. We compare an L1 loss to a Smooth L1 Loss in Table~\ref{table:loss}. The results show that the L1 loss is a bit better than Smooth L1 loss. It is because the L1 loss has a higher tolerance and thus is more robust to the noise. In contrast, the Smooth L1 loss is smoother but sensitive to outliers.

\noindent\textbf{Sensitivity Test.}
{ The default values of hyperparameters are set to $\lambda_{pix}$, $\lambda_{off}=1$, $\lambda_{s}=0.01$, $\lambda_{tri}=1$, and $\lambda_{con}=100$. Table~\ref{table:weight} shows the sensitivity test on different $\lambda_{s}$ and $\lambda_{con}$, which manifests that the accuracy slightly changes with different hyperparameter weights. Therefore, $\lambda_{s}$ and $\lambda_{con}$ are respectively set to $0.01$ and $100$ according to the results.}

\begin{table}
    \caption{Sensitivity test of different hyperparameters $\lambda_{s}$ and $\lambda_{con}$.}
    \label{tab:commands}
    \centering\sffamily
    \begin{tabular}{|l|c|c|c|c|}
    \hline
    \multirow{2}{*}{\bf{Weights}} & \multicolumn{3}{c|}{\bf{Wider Face}} & \multirow{2}{*}{\bf{MAFA}} \\ \cline{2-4}
    & Easy Set&   Medium Set&   Hard Set&   \\ \hline
    $\lambda_{s} = 0.01$ &  90.2\%& 85.4\%& 54.1\% & 91.5\%          \\ \hline
    $\lambda_{s} = 0.02$ & 89.9\%& 84.9\%& 53.4\%& 91.7\%           \\ \hline
    $\lambda_{s} = 0.2$ & 90.0\%& 85.5\%& 54.7\%&  90.6\%            \\ \hline
    $\lambda_{con} = 1$&   89.9\%&  85.3\%& 54.4\%& 91.5\%           \\ \hline
    $\lambda_{con} = 10$ &  89.9\%&  84.9\%& 53.4\%& 91.2\%          \\ \hline
    $\lambda_{con} = 100$ & 90.2\%& 85.4\%& 54.1\%& 91.5\%        \\
    \hline
    \end{tabular}
    \label{table:weight}
\end{table}

\section{Conclusions}\label{sec:conclusions}
To prevent the massive infections of the coronavirus and reduce the overloading of healthcare, we propose a new framework named "\verb|CenterFace|" to automate the monitoring of wearing face masks. "\verb|CenterFace|" uses the context attention module to enable the effective attention of the feed-forward convolution neural network by adapting their attention maps feature refinement. Moreover, we further propose an anchor-free detector with Triplet-Consistency Representation Learning by integrating the consistency loss and the triplet loss to deal with the small-scale training data and the similarity between masks and occlusions. Experimental results show that "\verb|CenterFace|" outperforms the other state-of-the-art methods, which is released as a public download to improve public health. In the future, we plan to study the problem of face recognition with masks and jointly consider the privacy issues. 

\begin{acks}
This work is supported in part by the Ministry of Science and Technology (MOST) of Taiwan under the grants MOST-109-2223-E-009-002-MY3, MOST-110-2634-F-007-015, MOST-109-2218-E-002-015, MOST-109-2221-E-009-114-MY3, MOST-110-2218-E-A49-018, MOST-109-2327-B-010-005, MOST-109-2221-E-009-097 and MOST-109-2221-E-001-015, and the Higher Education Sprout Project of the National Yang Ming Chiao Tung University and Ministry of Education (MOE), Taiwan.
\end{acks}

\bibliographystyle{ACM-Reference-Format}
\bibliography{sample-base}
\end{document}


\maketitle

\section{Implementation of Consistency Loss}
In this section, we introduce how we implement the consistency loss in detail. $\hat{\mathcal{Y}}$ and $\hat{\mathcal{Y}}'$ denote the predicted heatmap of the original image and the predicted heatmap of the horizontally-flipped images, respectively. The heatmaps predicted by the proposed model reveals the category and location information of the objects in the input images. To fully make the spatial and categorical information between the two prediction similar, for all the experiments, we utilize the loss function in Eq.~\ref{consistency_loss} as the $L_{con}$ to train the proposed model.
Moreover, we modify heatmap of the horizontally-flipped images before we calculate the loss function, since the heatmap of the horizontally-flipped images gives us the location of the object with horizontally-flipped. Therefore, we flip the heatmap of the horizontally-flipped images back to match the heatmap of the original image.
\begin{equation}\label{consistency_loss}
    L_{con} =\frac{1}{N}\sum_{i \in H'}\sum_{j \in W'}\mathcal{M}\sum_{c \in C}\left \|\hat{\mathcal{Y}}_{\mathbf{p}_n}-\mathcal{F}(\hat{\mathcal{Y}}'_{\mathbf{p}^\prime_n})\right \|^{2}.
\end{equation}
where $M$ is the boolean mask with size $H'\times W'$ adopted to filter out most backgrounds, where $H'$ and $W'$ are the spatial dimension of the heatmap. When the pixel is inside the area of the object, the pixel of the mask $M$ is 1; otherwise the pixel is set to 0. $\mathcal{F}$ is the flip operation to flip the heatmap. $N$ is the normalization term which is the summation of the mask, and C is the number of the categories of the dataset.